%% file: main_arxiv.tex
\documentclass[11pt,table]{article}
\pdfoutput=1

\usepackage[utf8]{inputenc} 
\usepackage[T1]{fontenc}    
\usepackage[in]{fullpage}
\usepackage[svgnames, table]{xcolor} 
\usepackage{xurl}
\usepackage{booktabs}       
\usepackage{amsfonts}       
\usepackage{nicefrac}       
\usepackage{microtype}      
\usepackage{amsmath,amssymb,amsthm}
\usepackage{mathtools}
\usepackage{mathrsfs}
\usepackage{thmtools}
\usepackage{thm-restate}
\usepackage{comment}
\usepackage{parskip}
\usepackage{siunitx}
\usepackage{etoolbox}
\usepackage{enumitem}
\usepackage{graphicx}
\usepackage{subcaption}
\usepackage[textsize=scriptsize,textwidth=2.2cm]{todonotes}
\usepackage{tabularx}
\usepackage[numbers,sort]{natbib}
\usepackage{hyperref}
\usepackage{longtable}
\usepackage{afterpage}
\usepackage{etoc}
\usepackage{makecell}
\usepackage{cleveref}
\usepackage{courier}
\usepackage{multirow}
\usepackage{pifont}
\newcommand{\cmark}{\ding{51}}%
\newcommand{\xmark}{\ding{55}}%
\newcolumntype{L}[1]{>{\raggedright\let\newline\\\arraybackslash\hspace{0pt}}m{#1}}
\newcolumntype{C}[1]{>{\centering\let\newline\\\arraybackslash\hspace{0pt}}m{#1}}
\newcolumntype{R}[1]{>{\raggedleft\let\newline\\\arraybackslash\hspace{0pt}}m{#1}}

\input{macros}

\makeatletter
\def\@maketitle{\newpage
 \null
 \vskip 2em
 \begin{center}%
  {\LARGE \@title \par}%
  \vskip 1.5em
    \def\And{%
    \end{tabular}\hfil\linebreak[0]\hfil%
    \begin{tabular}[t]{c}\rule{\z@}{24\p@}\ignorespaces%
    }
    \def\AND{%
    \end{tabular}\hfil\linebreak[4]\hfil%
    \begin{tabular}[t]{c}\rule{\z@}{24\p@}\ignorespaces%
    }
  {\large
   \lineskip .5em
    \begin{tabular}[t]{c}\rule{\z@}{24\p@}\@author\end{tabular}%
  }
  \vskip 1em
  {\large \@date}%
 \end{center}%
 \par
 \vskip 1.5em}
\makeatother

\title{Do Image Classifiers Generalize Across Time?}
\author{Vaishaal Shankar\thanks{Equal contribution} \\ UC Berkeley\and Achal Dave\samethanks \\ CMU\and Rebecca Roelofs\\ UC Berkeley \AND Deva Ramanan \\ CMU \and Benjamin Recht \\ UC Berkeley \and Ludwig Schmidt\\ UC Berkeley}

\date{}

\begin{document}

\maketitle
\input{abstract}

\input{introduction}

\input{naturaladversaries}

\input{experiments}
\input{related_work}

\input{conclusion}

\section*{Acknowledgements}
\input{acknowledgements}
\bibliographystyle{plainnat}
\bibliography{cites}{}

\newpage
\clearpage
\appendix
\input{appendix.tex}
\end{document}

%% file: macros.tex
\newtoggle{comments}
\togglefalse{comments}
\iftoggle{comments}{%
    \newcommand{\achal}[1]{{\todo[color=Orange!10]{Achal: #1}}}
    
    \newcommand{\becca}[1]{{\todo[color=Red!10]{Becca: #1}}}
    \newcommand{\deva}[1]{{\todo[color=Blue!10]{Deva: #1}}}
    \newcommand{\ludwig}[1]{\todo[color=Green!10]{Ludwig: #1}}
    \newcommand{\ludwiginline}[1]{\todo[inline,color=Green!10]{Ludwig: #1}}
    \newcommand{\vaishaal}[1]{{\todo[color=Purple!10]{Vaishaal: #1}}}
}{%
    \newcommand{\achal}[1]{}
    \newcommand{\becca}[1]{}
    \newcommand{\deva}[1]{}
    \newcommand{\ludwig}[1]{}
    \newcommand{\ludwiginline}[1]{}
    \newcommand{\vaishaal}[1]{}
}
\newcommand{\pmk}{PM-k}
\newcommand{\dataset}{\texttt{ImageNet-Vid-Robust}}
\newcommand{\datasetshlens}{\texttt{YTBB-Robust}}
\newcommand{\datasetorig}{ImageNet-Vid}
\newcommand{\datasetshlensorig}{Youtube-BB}

\newcommand{\ytbbreviewedpmk}{45,631}
\newcommand{\ytbbgoodpmk}{36,827}

\newcommand{\ytbbreviewedanchors}{2,467}
\newcommand{\ytbbgoodanchors}{2,030}

\newcommand{\ytbbcorrectedanchors}{834}
\newcommand{\ytbbcorrectedanchorsperc}{41\%}
\newcommand{\ytbbavgaddedlabels}{0.5}

\newcommand{\imvidreviewedpmk}{26,029}
\newcommand{\imvidgoodpmk}{21,070}

\newcommand{\imvidreviewedanchors}{1,314}
\newcommand{\imvidgoodanchors}{1,109}

\newcommand{\totalgoodanchors}{3,139}
\newcommand{\totalreviewedpmk}{71,660}
\newcommand{\totalgoodpmk}{57,897}

\newcommand{\mediandrop}{16}
\newcommand{\mediandropshlens}{10}
\newcommand{\accb}{$\text{acc}_{\text{orig}}$}
\newcommand{\acca}{$\text{acc}_{\text{pmk}}$}
\newcommand{\numclassifiers}{45}
\newcommand*\samethanks[1][\value{footnote}]{\footnotemark[#1]}
\newif\ifpreprint
\preprintfalse

\newcommand{\midsepremove}{\aboverulesep = 0mm \belowrulesep = 0mm}

\midsepremove

%% file: abstract.tex
\begin{abstract}
    We study the robustness of image classifiers to temporal perturbations derived from videos.
    As part of this study, we construct two datasets \footnote{Our datasets are available at this url: \url{https://modestyachts.github.io/robust-vid/}}, {\dataset} and {\datasetshlens}, containing a total of {\totalgoodpmk} images grouped into {\totalgoodanchors} sets of perceptually similar images.
    Our datasets were derived from {\datasetorig} and {\datasetshlensorig} respectively and thoroughly re-annotated by human experts for image similarity.
    We evaluate a diverse array of classifiers pre-trained on ImageNet and show a median classification accuracy drop of {\mediandrop} and {\mediandropshlens} on our two datasets.
    Additionally, we evaluate three detection models and show that natural perturbations induce both classification as well as localization errors, leading to a median drop in detection mAP of 14 points.
    Our analysis demonstrates that perturbations occurring naturally in videos pose a substantial and realistic challenge to deploying convolutional neural networks in environments that require both reliable and low-latency predictions.
\end{abstract}

%% file: introduction.tex
\section{Introduction}

Convolutional neural networks (CNNs) still exhibit many troubling failure modes.
At one extreme, $\ell_{p}$-adversarial examples cause large drops in accuracy for state-of-the-art models while relying only on visually imperceptible changes to the input image \citep{goodfellow2014explaining,biggio2017wild}.
However, this failure mode usually does not pose a problem outside a fully adversarial context because carefully crafted $\ell_p$-perturbations are unlikely to occur naturally in the real world.

To study more realistic failure modes, researchers have investigated benign image perturbations such as rotations \& translations, colorspace changes, and various image corruptions \citep{fawzi2015manitest,engstrom2017rotation,hosseini2018semantic,hendrycks2019benchmarking}.
However, it is still unclear whether these perturbations reflect the robustness challenges arising in real data since the perturbations also rely on synthetic image modifications.

Recent work has therefore turned to videos as a source of \emph{naturally occurring} perturbations of images \citep{Zheng_2016,azulay2018deep,gu2019using}.
In contrast to other failure modes, the perturbed images are taken from existing image data without further modifications to make the task more difficult.
As a result, robustness to such perturbations directly corresponds to performance improvements on real data.

However, it is currently unclear to what extent such video perturbations pose a robustness challenge.
\citet{azulay2018deep} and \citet{Zheng_2016} only provide anecdotal evidence from a small number of videos.
\citet{gu2019using} go beyond individual videos and utilize a large video dataset \citep{real2017youtube} in order to measure the effect of video perturbations more quantitatively.
In their evaluation, the best image classifiers lose about 3\% accuracy for video frames up to 0.3 seconds away.
However, the authors did not employ humans to review the frames in their videos.
Hence the accuracy drop could also be caused by significant changes in the video frames (e.g., due to fast camera or object motion).
Since the 3\% accuracy drop is small to begin with, it remains unclear whether video perturbations are a robustness challenge for current image classifiers.

\begin{figure}[t!]
    \centering
        \ifpreprint
        \includegraphics[width=\textwidth]{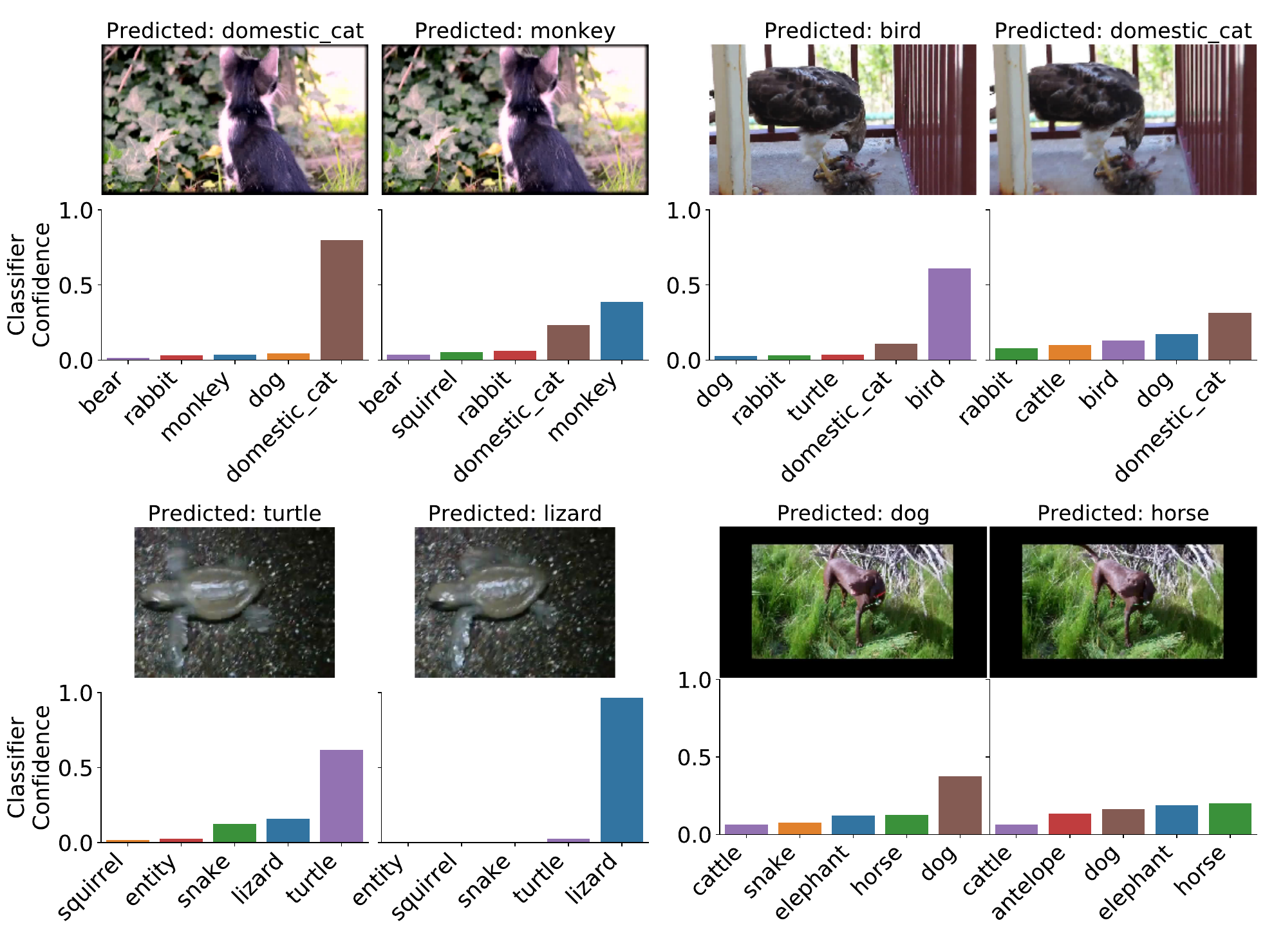}
        \else
        \includegraphics[width=\textwidth]{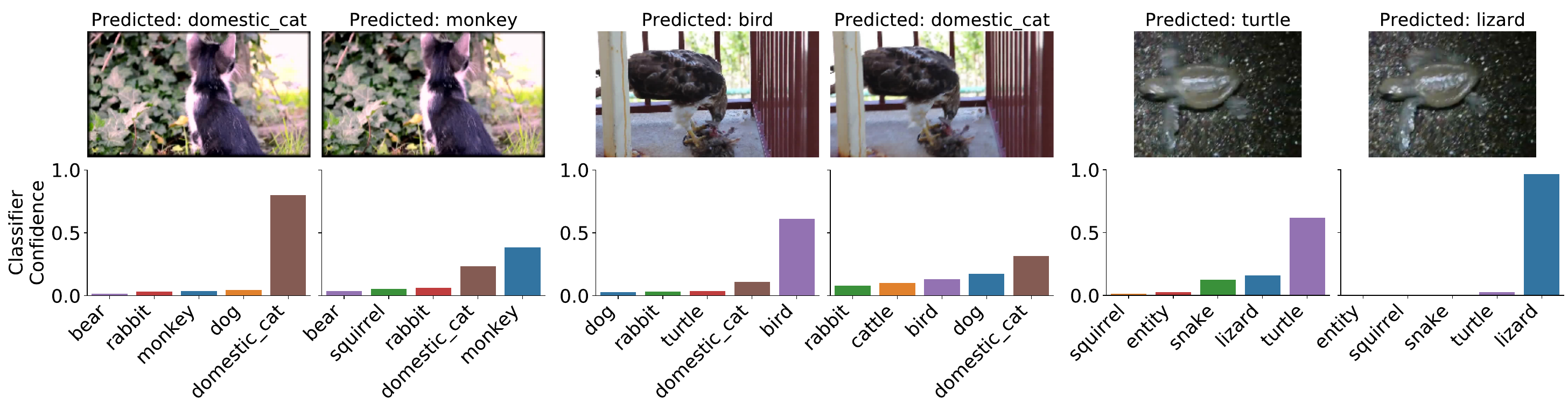}
        \fi
        \caption{\label{fig:natadv_example} Three examples of natural perturbations from nearby video frames and resulting classifier predictions from a ResNet-152 model fine-tuned on \datasetorig. While the images appear almost identical to the human eye, the classifier confidence changes substantially.
        }
       
\end{figure}

We address these issues by conducting a thorough evaluation of robustness to natural perturbations arising in videos.
As a cornerstone of our investigation, we introduce two test sets for evaluating model robustness: {\dataset} and {\datasetshlens}, carefully curated from the \datasetorig{} and \datasetshlensorig{} datasets, respectively \citep{ILSVRC15,real2017youtube}.
All images in the two datasets were screened by a set of expert labelers to ensure high annotation quality and minimize selection biases that arise when filtering a dataset with CNNs.
 To the best of our knowledge these are the first datasets of their kind, containing tens of thousands of images that are \emph{human reviewed} and grouped into thousands of perceptually similar sets. In total, our datasets contain {\totalgoodanchors} sets of temporally adjacent and visually similar images ({\totalgoodpmk} images total).

We then utilize these datasets to measure the robustness of current CNNs to small, naturally occurring perturbations.
Our testbed contains over {\numclassifiers} different models, varying both architecture and training methodology (adversarial training, data augmentation, etc.).
To better understand the drop in accuracy due to natural perturbations, we also introduce a robustness metric that is more stringent than those employed in prior work.
Under this metric, we find that natural perturbations from {\dataset} and {\datasetshlens} induce a median accuracy drop of {\mediandrop}\% and {\mediandropshlens}\% respectively for classification tasks and a median 14 point drop in mAP for detection tasks.\footnote{We only evaluated detection on {\dataset} as bounding-box annotations in {\datasetshlensorig} are available only at a temporal resolution of 1 frame-per-second and hence not dense enough for our evaluation.}
Even for the best-performing classification models, we observe an accuracy drop of 14\% for {\dataset} and 8\% for {\datasetshlens}.

Our results show that robustness to natural perturbations in videos is indeed a significant challenge for current CNNs.
As these models are increasingly deployed in safety-critical environments that require both high accuracy and low latency (e.g., autonomous vehicles), ensuring reliable predictions on \emph{every frame} of a video is an important direction for future work.

%% file: naturaladversaries.tex
\section{Constructing a test set for robustness}
{\dataset} and {\datasetshlens} are sourced from videos in the \datasetorig{} and \datasetshlensorig{} datasets \citep{ILSVRC15,real2017youtube}. 
All object classes in \datasetorig{} and \datasetshlensorig{} are from the WordNet hierarchy \citep{wordnet} and direct ancestors of ILSVRC-2012 classes.
Using the WordNet hierarchy, we construct a canonical mapping from ILSVRC-2012 classes to \datasetorig{} and \datasetshlensorig{} classes, which allows us to evaluate off-the-shelf ILSVRC-2012 models on \dataset{} and \datasetshlens{}. 
We provide more background on the source datsets in Appendix ~\ref{sec:sourcedatasets}.

\subsection{Constructing \dataset{} and \datasetshlens{}}
\label{sec:construct}
\begin{figure}[t!]
    \includegraphics[width=\textwidth]{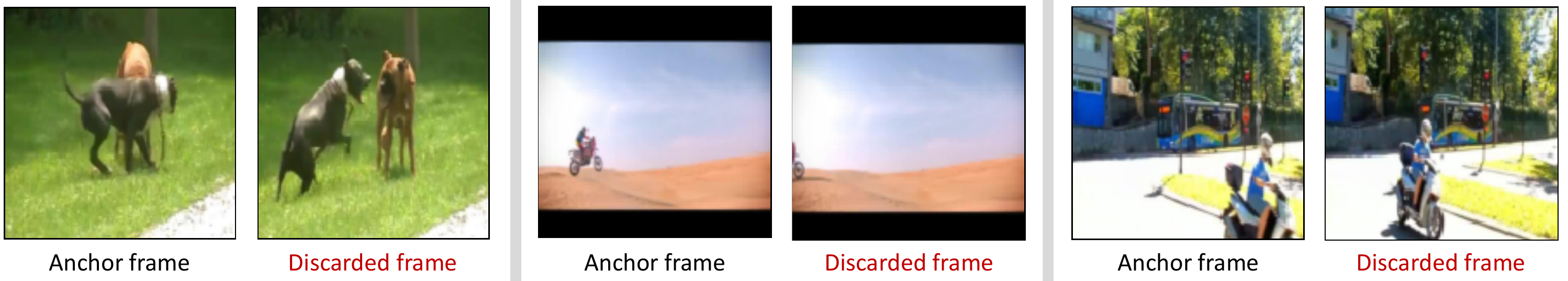}
    \caption{\label{fig:dissim_example} Temporally adjacent frames may not be visually similar. We show three randomly sampled frame pairs where the nearby frame was marked as ``dissimilar'' to the anchor frame during human review and then discarded from our dataset.}
\end{figure}

Next, we describe how we extracted sets of naturally perturbed frames from {\datasetorig} and {\datasetshlensorig} to create {\dataset} and {\datasetshlens}.
A straightforward approach would be to select a set of anchor frames and use temporally adjacent frames in the video with the assumption that such frames contain only small perturbations from the anchor.
However, as Figure ~\ref{fig:dissim_example} illustrates, this assumption is frequently violated, especially due to fast camera or object motion. 

Instead, we first collect \textit{preliminary} datasets of natural perturbations following the same approach, and then manually review each of the frame sets. 
For each video, we randomly sample an anchor frame and take $k=10$ frames before and after the anchor frame as candidate perturbation images.\footnote{For \datasetshlens{} we use a subset of the anchor frames used by \citet{gu2019using}.}
This results in two datasets containing one anchor frame each from {\totalgoodanchors} videos, with approximately 20 candidate perturbation per anchor frame.\footnote{Anchor frames near the start or end of the video may have less than 20 candidate frames.}

Next, we curate the dataset with the help of four expert human annotators. The goal of the curation step is to ensure that each anchor frame and its nearby frames are correctly labeled with the same ground truth class, and that the anchor frame and the nearby frames are visually similar.

\textbf{Denser labels for \datasetshlensorig.} As \datasetshlensorig{} contains only a single category label per frame at 1 frame per second, annotators first inspected each anchor frame individually and added any missing labels.
In total, annotators corrected the labels for \ytbbcorrectedanchors{} frames, adding an average of \ytbbavgaddedlabels{} labels per anchor frame.
These labels are then propagated to nearby, unlabeled frames at the native frame rate and verified in the next step.
\datasetorig{} densely labels all classes per frame, so we skipped this step for this dataset.

\textbf{Frame pairs review.}
Next, for each pair of anchor and nearby frames, a human annotates (i) whether the pair is correctly labeled in the dataset, and (ii) whether the pair is similar.
We took several steps to mitigate the subjectivity of this task and ensure high annotation quality.
First, we trained reviewers to mark frames as dissimilar if the scene undergoes any of the following transformations: significant motion, significant background change, or significant blur change.
We asked reviewers to mark each dissimilar frame with one of these transformations, or ``other'', and to mark a pair of images as dissimilar if a distinctive feature of the object is only visible in one of the two frames (such as the face of a dog).
If an annotator was unsure about the correct label, she could mark the pair as ``unsure''.
Second, we present only a single pair of frames at a time to reviewers because presenting videos or groups of frames could cause them to miss large changes due to the phenomenon of change blindness \citep{pashler1988familiarity}.

\textbf{Verification.} In the previous stage, all annotators were given identical labeling instructions and individually reviewed a total of \totalreviewedpmk{} image pairs. To increase consistency in annotation, annotators jointly reviewed all frames marked as dissimilar, incorrectly labeled, or ``unsure''.
A frame was only considered similar to its anchor if a strict majority of the annotators marked the pair as such. 

After the reviewing was complete, we discarded all anchor frames and candidate perturbations that annotators marked as dissimilar or incorrectly labeled. 
The final datasets contain a combined total of \totalgoodanchors{} anchor frames with a median of $20$ similar frames each.

\begin{table}
   \centering
   \begin{tabular}{cccC{3cm}}  
       \toprule
       & & \dataset & \datasetshlens \\\midrule
       \multirow{3}{*}{Anchor frames} & Reviewed  & \imvidreviewedanchors           & \ytbbreviewedanchors \\
        & Accepted  & \imvidgoodanchors{}\rlap{ (84\%)} & \ytbbgoodanchors{}\rlap{ (82\%)} \\
        & Labels updated & -       & \ytbbcorrectedanchors{}\rlap{    (41\%)}  \\
       \midrule
       \multirow{2}{*}{Frame pairs} &
          Reviewed        & \imvidreviewedpmk       & \ytbbreviewedpmk \\
        & Accepted        & \imvidgoodpmk{}\rlap{ (81\%)} & \ytbbgoodpmk{}\rlap{ (81\%)} \\
       \bottomrule
   \end{tabular}
  
   \caption{ \label{tab:dataset-statistics} Dataset statistics of \dataset{} and \datasetshlens. For \datasetshlens, we updated the labels from for \ytbbcorrectedanchorsperc{} (\ytbbcorrectedanchors) of the accepted anchors due to incomplete labels in \datasetshlensorig.}
\end{table}

\subsection{The \texttt{pm-k} evaluation metric}
\label{sec:metric}
Given the datasets introduced above, we propose a metric to measure a model's robustness to natural perturbations. In particular, let $A = \{a_1, ..., a_n\}$ be the set of valid anchor frames in our dataset. Let $Y = \{y_{1}, ..., y_{n}\}$ be the set of labels for $A$.
 We let $\mathcal{N}_{k}(a_i)$ be the set of frames marked as similar to anchor frame $a_i$. 
 In our setting, $\mathcal{N}_{k}$ is a subset of the $2k$ temporally adjacent frames (plus/minus k frames from the anchor). 

\paragraph{Classification.}
The standard classification accuracy on the anchor frame is
$\text{acc}_{\text{orig}} = 1 - \frac{1}{N} \sum_{i=1}^{N} \mathcal{L}_{0/1}(f(a_{i}), y_{i})$, where $\mathcal{L}_{0/1}$ is the standard 0-1 loss function.  
We define the \texttt{pm-k} analog of accuracy as
\begin{align}
    \text{acc}_{\text{pmk}} \; = \; 1 - \frac{1}{N} \sum_{i=1}^{N} \max_{b \in \mathcal{N}_{k}(a_{i})}
    \mathcal{L}_{0/1}(f(b), y_{i}) \; , \label{acc:adv}
\end{align}
which corresponds to picking the worst frame from each set $\mathcal{N}_{k}(a_{i})$ before computing accuracy. We note the similarity of the \texttt{pm-k} metric to  standard $\ell_{p}$ adversarial robustness. If we let $\mathcal{N}_{k}(a_{i})$ be the set of \textit{all} images within an $\ell_{p}$ ball of radius $\epsilon$  around $a_{i}$, then the two notions of robustness are identical.

\paragraph{Detection.} The standard metric for detection is mean average precision (mAP) of the predictions at a fixed intersection-over-union (IoU) threshold~\cite{lin2014microsoft}. We define the \texttt{pm-k} metric analogous to that for classification: We replace each anchor frame with the nearest frame that minimizes the average precision (AP, averaged over recall thresholds) of the predictions, and compute \texttt{pm-k} as the mAP on these worst-case neighboring frames.

%

%% file: experiments.tex
\section{Main results}
\label{sec:experiments}

\begin{figure}[b!]
    \centering
    \begin{subfigure}{0.4\textwidth}
    \includegraphics[width=\textwidth]{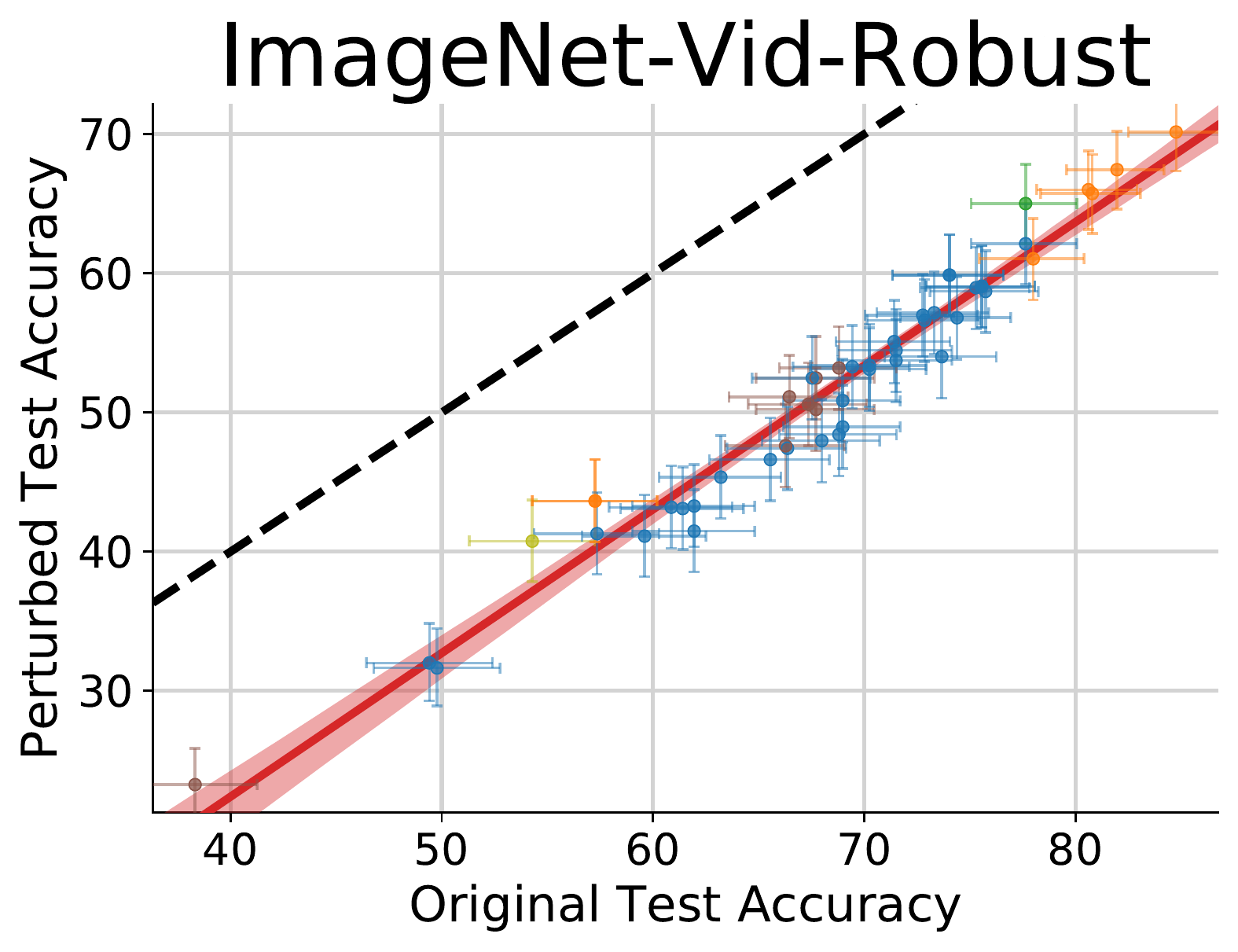}
    \end{subfigure}
    \begin{subfigure}{0.4\textwidth}
    \includegraphics[width=\textwidth]{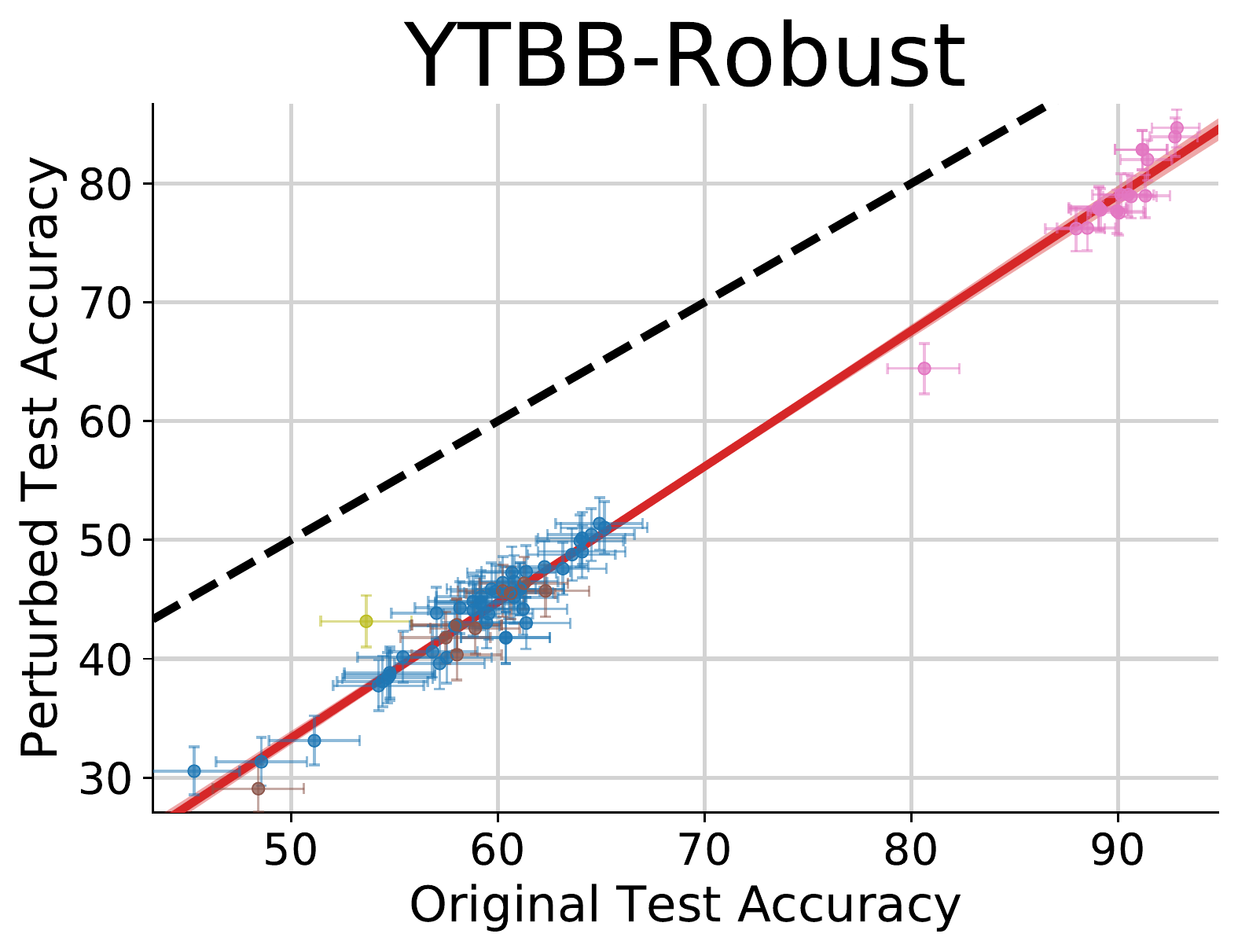}
    \end{subfigure}
    \begin{subfigure}{0.8\textwidth}
    \includegraphics[width=\textwidth]{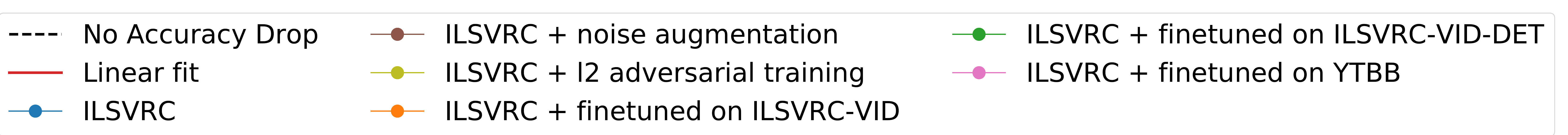}
    \end{subfigure}
    \caption{Model accuracy on original vs. perturbed images. 
    Each data point corresponds to one model in our testbed (shown with 95\% Clopper-Pearson confidence intervals). Each perturbed frame was taken from a ten frame neighborhood of the original frame (approximately 0.3 seconds). All frames were reviewed by humans to confirm visual similarity to the original frames. \label{fig:pm10}}

\end{figure}

We evaluate a testbed of \numclassifiers{} classification and three detection models on {\dataset} and {\datasetshlens}.
We first discuss the various types of classification models evaluated with the \texttt{pm-k} classification metric. 
Second, we evaluate the performance of detection models on {\dataset} using use the bounding box annotations inherited from \datasetorig{} and using a variant of the \texttt{pm-k} metric for detection.
We then analyze the errors made on the detection adversarial examples to 
isolate the effects of \emph{localization} errors vs. \emph{classification} errors.

\subsection{Classification}
\label{sec:experiments-classification}
The classification robustness metric is {\acca} defined in \Cref{acc:adv}. For frames with multiple labels, we count a prediction as correct if the model predicts \emph{any} of the correct classes for a frame.
In \Cref{fig:pm10}, we plot the benign accuracy, {\accb}, versus the robust
accuracy, {\acca}, for all classification models in our test bed and find
that the relationship between {\accb} and {\acca} is approximately linear.
This relationship indicates that improvements in the benign accuracy do
result in improvements in the worst-case accuracy, but do not suffice to
resolve the accuracy drop due to natural perturbations.

Our test bed consists of five model types with increasing levels of supervision.
We present results for representative models from each model type in \Cref{model_type_table} and defer the full classification results table to Appendix \ref{app:table}.
\begin{table}
    \centering
    \caption{Accuracies of five different model types and the best performing model (shown with 95\% Clopper-Pearson confidence intervals). $\Delta$ denotes accuracy drop between evaluation on anchor frame (\texttt{pm-0}) and worst frame in similarity set (\texttt{pm-10}). The model architecture is ResNet-50 unless noted otherwise. `FT' denotes `fine-tuning.' See \Cref{sec:experiments-classification} for details. }
\label{model_type_table}
\rowcolors{2}{white}{gray!25}
\begin{tabular}{lccc}
    \rowcolor{white!50}
    \toprule
    Model Type  &  \thead{Accuracy \\ Original}&  \thead{Accuracy\\ Perturbed} & $\Delta$ \\
    \midrule
    \rowcolor{white!50}
    \multicolumn{4}{c}{\dataset} \\\midrule
    Trained on ILSVRC
        & 67.5 {\footnotesize \textcolor{gray}{[64.7, 70.3]}}
        & 52.5 {\footnotesize \textcolor{gray}{[49.5, 55.5]}}
        & 15.0 \\
    + Noise Augmentation
        & 68.8 {\footnotesize \textcolor{gray}{[66.0, 71.5]}}
        & 53.2 {\footnotesize \textcolor{gray}{[50.2, 56.2]}}
        & 15.6 \\
    + $\ell_{\infty}$ robustness (ResNext-101)
        & 54.3 {\footnotesize \textcolor{gray}{[51.3, 57.2]}}
        & 40.8 {\footnotesize \textcolor{gray}{[39.0, 43.7]}}
        & 12.4 \\
    + FT on \datasetorig{}
        & 80.8 {\footnotesize \textcolor{gray}{[78.3, 83.1]}}
        & 65.7 {\footnotesize \textcolor{gray}{[62.9, 68.5]}}
        & 15.1\\
    + FT on \datasetorig{} (ResNet-152) 
        &  84.8 {\footnotesize \textcolor{gray}{[82.5, 86.8]}}
        & 70.2 {\footnotesize \textcolor{gray}{[67.4, 72.8]}}
        & 14.6\\ 
    + FT on \datasetorig-Det
        & 77.6 {\footnotesize \textcolor{gray}{[75.1, 80.0]}}
        & 65.4 {\footnotesize \textcolor{gray}{[62.5, 68.1]}}
        & 12.3\\
    \midrule
    \rowcolor{white!50}
    \multicolumn{4}{c}{\datasetshlens} \\\midrule
    Trained on ILSVRC
        & 57.0 {\footnotesize \textcolor{gray}{[54.9, 59.2]}}
        & 43.8 {\footnotesize \textcolor{gray}{[41.7, 46.0]}}
        & 13.2\\\
    + Noise Augmentation
        & 62.3 {\footnotesize \textcolor{gray}{[60.2, 64.4]}}
        & 45.7 {\footnotesize \textcolor{gray}{[43.5, 47.9]}}
        & 16.6 \\
    + $\ell_{\infty}$ robustness (ResNext-101)
        & 53.6 {\footnotesize \textcolor{gray}{[51.4, 55.8]}}
        & 43.2 {\footnotesize \textcolor{gray}{[41.0, 45.3]}}
        & 10.4 \\
    + FT on \datasetshlensorig{}
        & 91.4 {\footnotesize \textcolor{gray}{[90.1, 92.6]}}
        & 82.0 {\footnotesize \textcolor{gray}{[80.3, 83.7]}}
        & 9.4 \\
    + FT on \datasetshlensorig{} (ResNet-152)
        & 92.9 {\footnotesize \textcolor{gray}{[91.6, 93.9]}}
        & 84.7 {\footnotesize \textcolor{gray}{[83.0, 86.2]}}
        & 8.2 
    \end{tabular}
\end{table}
\parskip=1.5pt
\paragraph{ILSVRC Trained}
The WordNet hierarchy enables us to repurpose models trained for the 1,000 class ILSVRC-2012 dataset on {\dataset} and {\datasetshlens} (see \Cref{sec:dataset}).
We evaluate a wide array of ILSVRC-2012 models (available from \cite{cadene}) against our natural perturbations.
Since these datasets present a substantial distribution shift from the original ILSVRC-2012 validation set, we expect the \emph{benign} accuracy {\accb} to be lower than the comparable accuracy on the ILSVRC-2012 validation set. 
However, our main interest here is in the \emph{difference} between the original and perturbed accuracies {\accb} - {\acca}. A small drop in accuracy would indicate that the model is robust to small changes that occur naturally in videos. Instead, we find significant median drops of 15.0\% and 13.2\% in accuracy on our two datasets, indicating sensitivity to such changes.

\paragraph{Noise augmentation}
One hypothesis for the accuracy drop from original to perturbed accuracy is that subtle artifacts and corruptions introduced
by video compression schemes could degrade performance when evaluating on these corrupted frames. The worst-case nature of the \texttt{pm-k} metric could then be focusing on these corrupted frames.
One model for these corruptions are the perturbations introduced in \cite{hendrycks2019benchmarking}.
To test this hypothesis, we evaluate models augmented with a subset of the perturbations (exactly one of: Gaussian noise, Gaussian blur, shot noise, contrast change, impulse noise, or JPEG compression).
We found that these augmentation schemes did not improve robustness against our perturbations substantially, and still result in a median accuracy drop of 15.6\% and 16.6\% on the two datasets.

\paragraph{$\ell_{\infty}$-robustness.}
We evaluate the model from ~\cite{xie2018feature}, which currently performs best against $\ell_{\infty}$-attacks on ImageNet.
We find that this model has a smaller accuracy drop than the two aforementioned model types on both datasets. However, the robust model achieves substantially lower original and perturbed accuracy than either of the two model types above, and the robustness gain is modest (3\% compared to models of similar benign accuracy).

\paragraph{Fine-tuning on video frames.}
To adapt to the new class vocabulary and the video domain, we fine-tune several network
architectures on the {\datasetorig} and {\datasetshlensorig} training sets.
For \datasetshlensorig, we train on the anchor frames used for training in \cite{gu2019using}, and for
{\datasetorig} we use all frames in the training set.
We provide hyperparameters for all models in Appendix \ref{app:expdetails}.

The resulting models significantly improve in accuracy over their ILSVRC pre-trained counterparts (e.g., 13\% on \dataset{} and 34\% on \datasetshlens{} for ResNet-50).
This improvement in accuracy results in a modest improvement in the accuracy drop for \datasetshlens{}, but a finetuned ResNet-50 still suffers from a substantial 9.4\% drop.
On \dataset{}, there is almost no change in the accuracy drop from 15.0\% to 15.1\%.

\paragraph{Fine-tuning for detection on video frames.}
We further analyze whether additional supervision in the form of bounding box annotations improves robustness. 
To this end, we train the Faster R-CNN \textit{detection} model~\cite{ren2015faster} with a ResNet-50 backbone on \datasetorig. Following standard practice, the detection backbone is pre-trained on ILSVRC-2012.
To evaluate this detector for classification, we assign the class with the most confident bounding box as label to the image.
We find that this transformation reduces accuracy compared to the model trained for classification (77.6\% vs.\ 80.8\%).
While there is a slight reduction in the accuracy drop caused by natural perturbations, the reduction is well within the error bars for this test set. We leave an in-depth investigation of additional supervision to induce robustness for future work.

\subsection{Detection}
We further study the impact of natural perturbations on object detection. Specifically, we report results for two related tasks: object localization and detection.
Object detection is the standard computer vision task of correctly classifying an object and finding the coordinates of a tight bounding box containing the object.
``Object localization'', meanwhile, refers to only the subtask of finding the bounding box, \textit{without} attempting to correctly classify the object.

\begin{figure}[ht!]
\includegraphics[width=\linewidth]{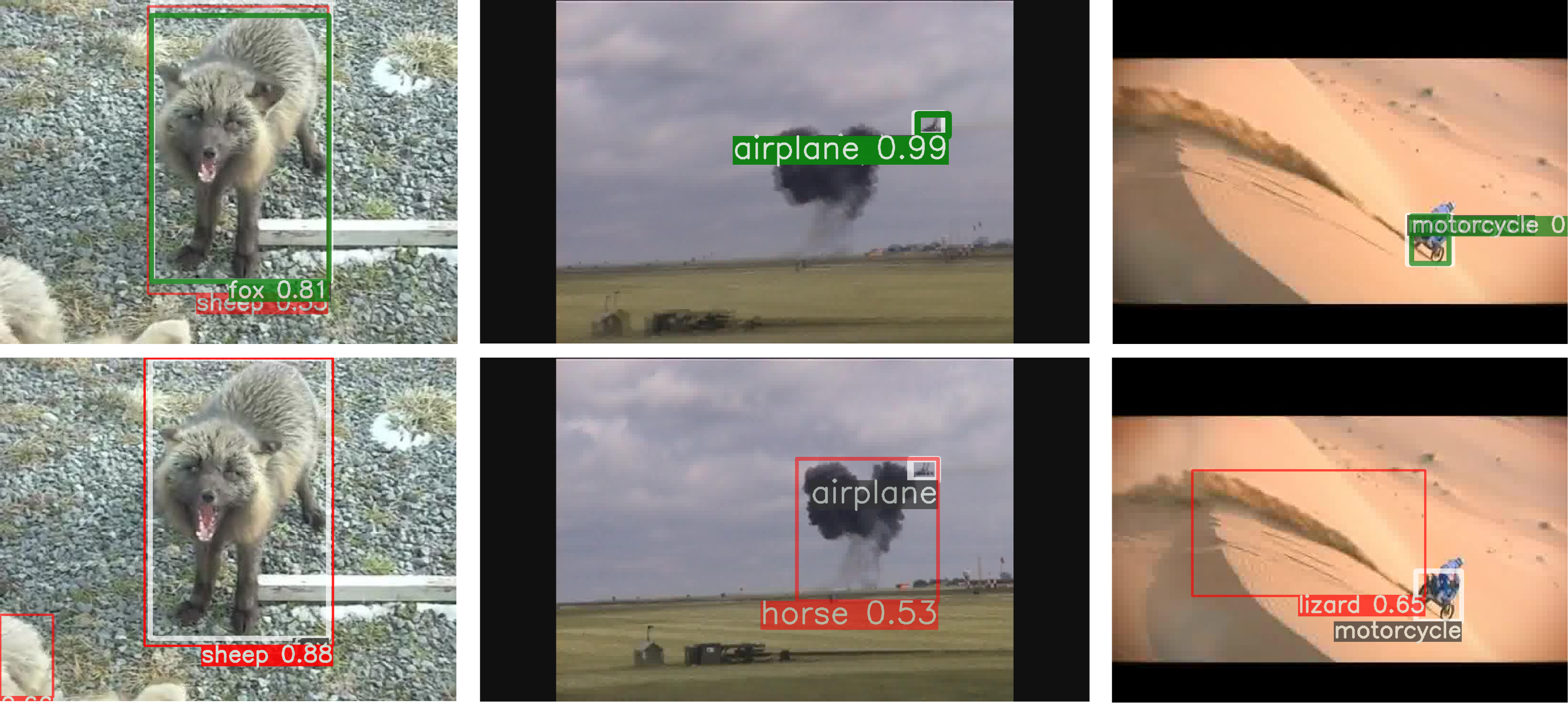}
\caption{Naturally perturbed examples for detection. Red boxes indicate false positives; green boxes indicate true positives; white boxes are ground truth. Classification errors are common failures, such as the fox on the left, which is classified correctly in the anchor frame, and misclassified as a sheep in a nearby frame. However, detection models also have \textit{localization} errors, where the object of interest is not correctly localized in addition to being misclassified, such as the airplane (middle) and the motorcycle (right). All visualizations show predictions with confidence greater than 0.5.}
\label{fig:detection-failures}
\end{figure}

We provide our results on \dataset{}, which contains dense bounding box labels unlike \datasetshlensorig{}, which only labels boxes at 1 frame per second. 
We use the popular Faster R-CNN~\cite{ren2015faster} and R-FCN~\cite{dai2016r,xiao2018video} architectures for object detection and localization and report results in \Cref{tab:detection-results}. 
For the R-FCN architecture, we use the model from~\cite{xiao2018video}\footnote{This model was originally trained on the 2015 subset of \datasetorig. We evaluated this model on the 2015 validation set because the method requires access to pre-computed bounding box proposals which are available only for the 2015 subset of \datasetorig.}.
We first note the significant drop in mAP of 12 to 15 points for object detection due to perturbed frames for both the Faster R-CNN and R-FCN architectures. Next, we show that localization is indeed easier than detection, as the mAP is higher for localization than for detection (e.g., $76.6$ vs $62.8$ for Faster R-CNN with a ResNet-50 backbone). Perhaps surprisingly, however, switching to the localization task does \textit{not} improve the drop between original and perturbed frames, indicating that natural perturbations induce both classification and localization errors. We show examples of detection failures in \Cref{fig:detection-failures}.

\begin{table}[bt!]
\centering
\caption{Detection and localization mAP for two Faster R-CNN backbones. Both detection and localization suffer from significant drops in mAP due to the perturbations. (* indicates that the model has trained on the ILSVRC Det and VID 2015 datasets and evaluated on the 2015 subset of ILSVRC-VID 2017.)}
\rowcolors{2}{white}{gray!25}
\begin{tabular}{clccc}
    \rowcolor{white!50}
    \toprule
    Task & Model &\thead{mAP\\ Original} & \thead{mAP\\ Perturbed} & \thead{mAP\\ $\Delta$} \\
    \midrule
    \cellcolor{white!50}
        & FRCNN, ResNet 50   & 62.8 & 48.8 & 14.0 \\
    \cellcolor{white!50}
        & FRCNN, ResNet 101  & 63.1 & 50.6 & 12.5 \\
    \cellcolor{white!50}
    \multirow{-3}{*}{Detection}
        & R-FCN, ResNet 101 \cite{xiao2018video}* & 79.4\rlap{*} & 63.7\rlap{*} & 15.7\rlap{*} \\
    \midrule
    \cellcolor{white!50}
        & FRCNN, ResNet 50 & 76.6 & 64.2 & 12.4 \\
    \cellcolor{white!50}
        & FRCNN, ResNet 101 & 77.8 & 66.3 & 11.5 \\
    \cellcolor{white!50}
    \multirow{-3}{*}{Localization}
        & R-FCN, ResNet 101* & 80.9\rlap{*} & 70.3\rlap{*} & 10.6\rlap{*} \\
    \bottomrule
\end{tabular}
\label{tab:detection-results}
\end{table}

\subsection{Impact of dataset review}
We analyze the impact of our human review, described in \Cref{sec:construct}, on the classifiers in our testbed. First, we compare the original and perturbed accuracies of a representative classifier (ResNet-152 finetuned) with and without review in \Cref{tab:accuracies_without_review}. Our review improves the original accuracy by 3 to 4\% by discarding mislabeled or blurry anchor frames, and improves perturbed accuracy by 5 to 6\% by discarding pairs of dissimilar frames. Our review reduces the accuracy drop by 1.8\% on \dataset{} and 1.1\% on \datasetshlens{}. These results indicate that the changes in model predictions are indeed due to a lack of robustness, rather than due to significant differences between adjacent frames.

\begin{figure}[ht!]
\includegraphics[width=0.49\linewidth]{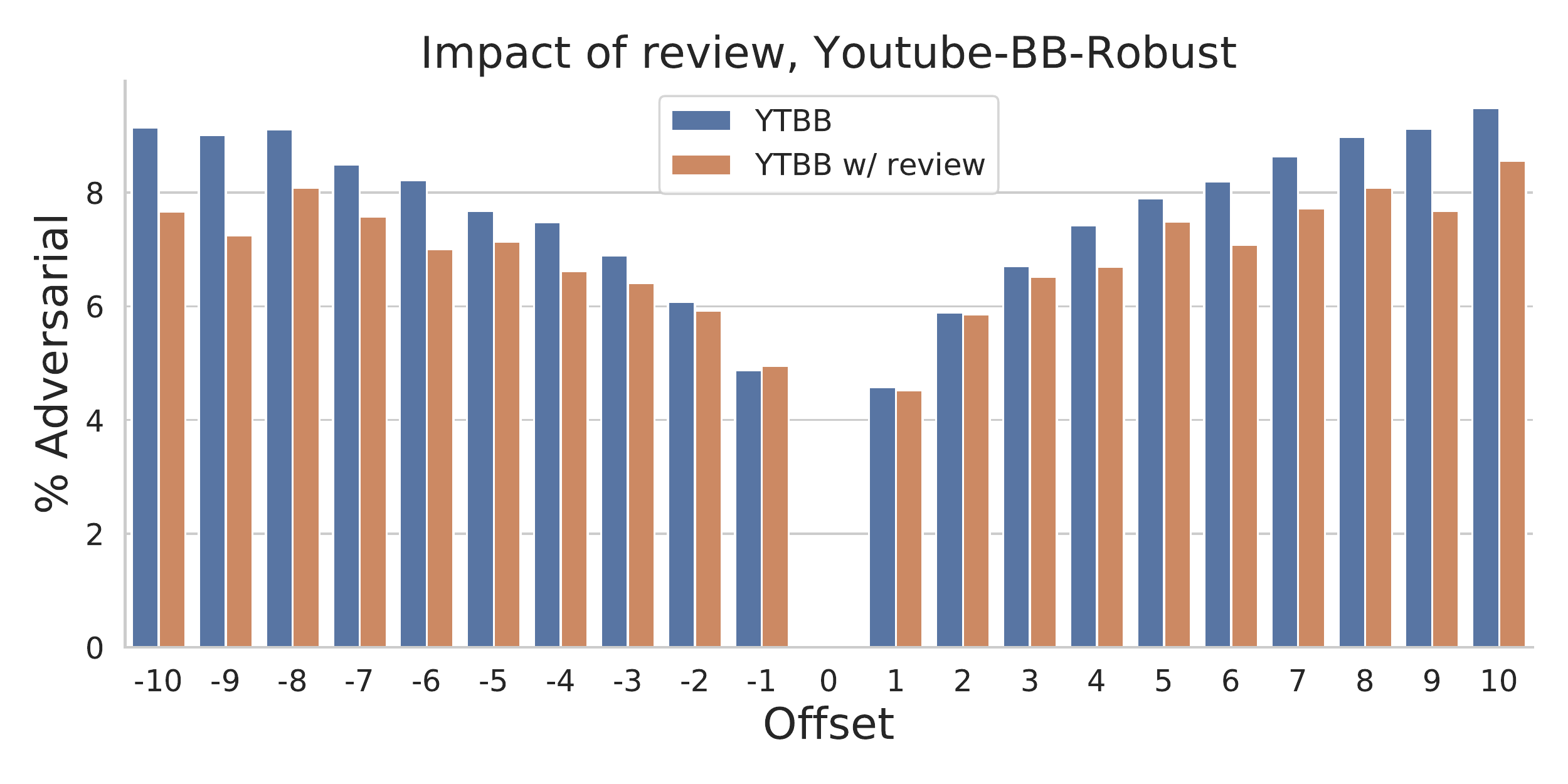}
\includegraphics[width=0.49\linewidth]{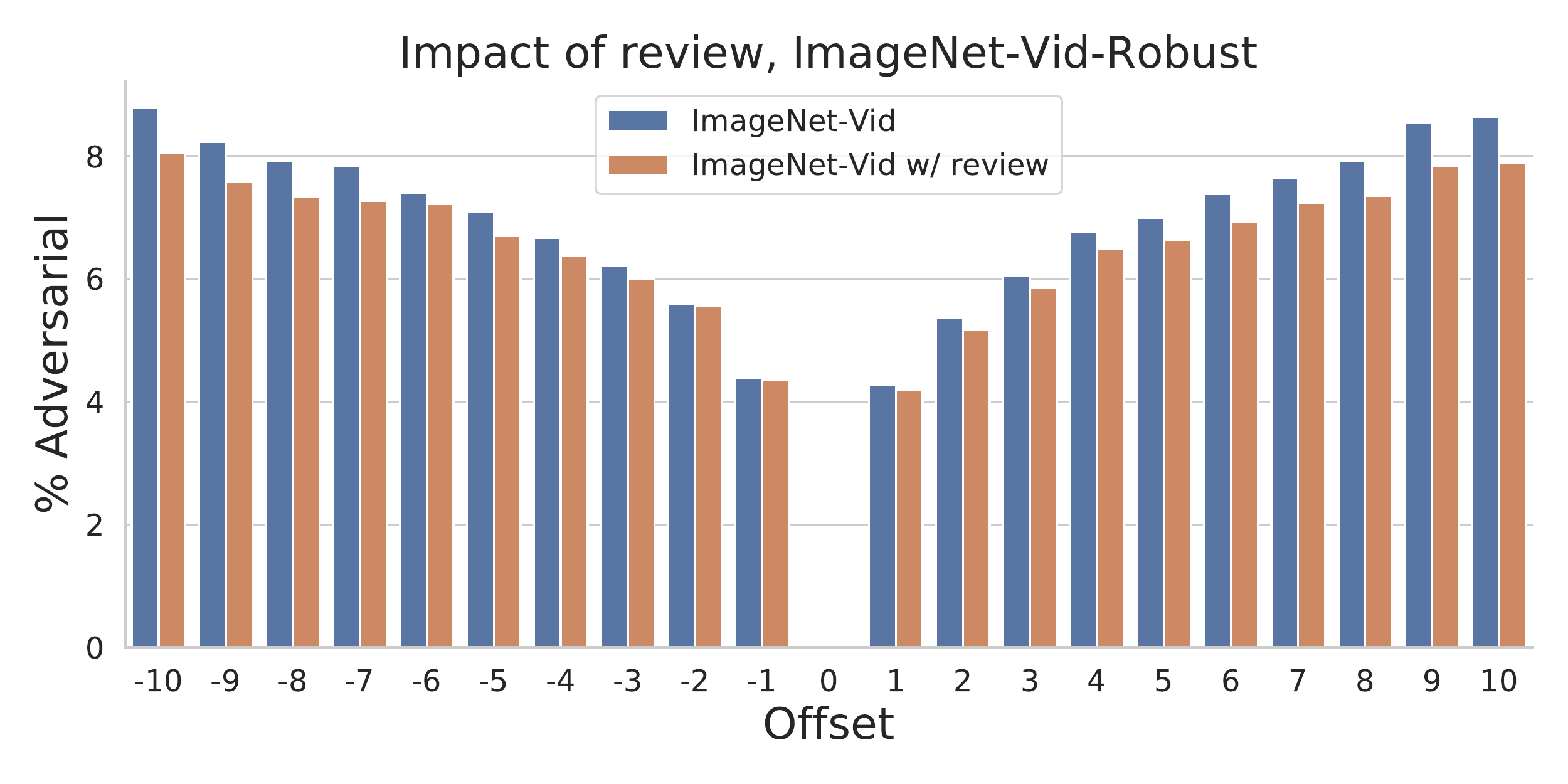}
\caption{We plot the fraction of times each offset caused an error, across all evaluated models, for frames with and without review. Frames further away more frequently cause classifiers to misfire. Our review process reduces the number of errors, especially for frames further in time, by removing dissimilar frames.}
\label{fig:histogram_errors}
\end{figure}

To further analyze the impact of our review on model errors, we plot how frequently each offset distance from the anchor frame results in a model error across all model types in \Cref{fig:histogram_errors}. For both datasets, larger offsets (indicating pairs of frames further apart in time) lead to more frequent model errors. Our review reduces the fraction of errors across offsets,  especially for large offsets, which are more likely to display large changes from the anchor frame.

\begin{table}[h!]
    \centering
    \caption{Impact of human review on original and perturbed accuracies for \dataset{} and \datasetshlens{}, numbers come from a ResNet-152 fine-tuned on \datasetorig{} and \datasetshlensorig{}, respectively.}
    \begin{tabular}{lcccccc}
        \toprule
                                    &           & \multicolumn{2}{c}{Accuracy} \\
                                    & Reviewed  & Original & Perturbed & $\Delta$ \\\midrule
        \multirow{2}{*}{\dataset{}} & \xmark    & 80.3 & 64.1 & 16.2 \\
                                    & \cmark    & 84.8 & 70.2 & 14.4 \\\midrule
        \multirow{2}{*}{\datasetshlens{}} & \xmark & 88.1 & 78.1 & 10.0 \\
                                          & \cmark & 92.9 & 84.7 & 8.9 \\
        \bottomrule
    \end{tabular}
    \label{tab:accuracies_without_review}
\end{table}

%% file: related_work.tex
\section{Related work}
\paragraph{Adversarial examples.}
Various forms of adversarial examples have been studied, the majority of research focuses on $\ell_{p}$ robustness \cite{goodfellow2014explaining,biggio2017wild}. 
However, it is unclear whether adversarial examples pose a problem for robustness outside
of a truly worst case context. It is an open question whether perfect robustness against a $\ell_{p}$
adversary will induce robustness to realistic image distortions such as those studied in this paper.
Recent work has proposed less adversarial image modifications such as small rotations \& translations \cite{engstrom2017rotation,azulay2018deep,fawzi2015manitest,kanbak2017geometric}, hue and color changes \cite{hosseini2018semantic}, image stylization \cite{geirhos2018imagenet} and synthetic image corruptions such as Gaussian blur and JPEG compression \cite{hendrycks2019benchmarking, geirhos2018generalisation}.
Even though the above examples are more realistic than the $\ell_p$ model, they still synthetically modify the input images to generate perturbed versions.
In contrast, our work performs no synthetic modification and instead uses images that naturally occur in videos.
\paragraph{Utilizing videos to study robustness.}
In work concurrent to ours, \citet{gu2019using} exploit the temporal structure in videos to study robustness. 
However, their experiments suggest a substantially smaller drop in classification accuracy. 
The primary reason for this is a less stringent metric used in \cite{gu2019using}. 
By contrast, our {\pmk} metric is inspired by the ``worst-of-k'' metric used in prior work~\cite{engstrom2017rotation}, highlighting the sensitivity of models to natural perturbations.  
In Appendix \ref{sec:shlens_compare} we study the differences between the two metrics in more detail.
Furthermore, the lack of human review and the high label error-rate we discovered in \datasetshlensorig (Table \ref{tab:dataset-statistics}) presents 
a potentially troubling confounding factor that we resolve in our work. 
\paragraph{Distribution shift.} 
Small, benign changes in the test distribution are often referred to as \textit{distribution shift}.
\citet{recht2019imagenet} explore this phenomenon by constructing new test sets for CIFAR-10 and ImageNet and observe substantial performance drops for a large suite of models on the newly constructed test sets. 
Similar to our Figure \ref{fig:pm10}, the relationship between their original and new test set accuracies is also approximately linear.
However, the images in their test set bear little visual similarity to images in the original test set, while all of our failure cases in {\dataset} and {\datasetshlens} are on perceptually similar images. In a similar vein of study, \cite{torralba2011unbiased} studies distribution shift \textit{across} different computer vision data sets such as Caltech-101, PASCAL, and ImageNet.
\paragraph{Temporal Consistency in Computer Vision} 
A common issue when applying image based models to videos is \textit{flickering}, where object detectors spuriously produce false-positives or false-negatives in isolated frames or groups of frames. 
\cite{jin2018unsupervised} explicitly identify such failures and use a technique reminiscent of adversarially robust training to improve image-based models. \ludwig{Is there code we can run for this?}\achal{They have some code for training their thing, but no shared models. Could try as baseline for when we try to fix this problem :)}
A similar line of work focuses on improving object detection in videos as objects become occluded or move quickly
\cite{kang2017object,feichtenhofer2017detect,zhu2017flow,xiao2018video}. 
The focus in this line of work has generally been on improving object detection when objects transform in a way that makes recognition difficult from a single frame, such as fast motion or occlusion. 
In this work, we document a broader set of failure cases for image-based classifiers and detectors and show that failures occur when the neighboring frames are imperceptibly different.

%% file: conclusion.tex
\section{Conclusion}
 

Our study quantifies the sensitivity of image classifiers
to naturally occuring temporal perturbations. 
These perturbations cause significant drops in
accuracy for a wide range of models in both classification and detection. Our work
on analyzing this failure mode opens multiple avenues for future research:
\paragraph{Building more robust models.} Our {\dataset} and {\datasetshlens} datasets provide a standard
measure for robustness that can be used to evaluate to any classification or detection
model. In~\Cref{model_type_table}, we evaluated several commonly used
models and found that all of them suffer from substantial accuracy drops due to natural
perturbations. In particular, we found that model improvements with respect to artificial perturbations (such as
image corruptions or $\ell_{\infty}$ adversaries) induce at best modest improvements in robustness.
We hope that our standardized datasets and evaluation metric will enable future work to
quantify improvements in natural robustness directly.
\paragraph{Further natural perturbations.} Videos provide a straightforward
method for collecting natural perturbations of images, enabling the study of
realistic forms of robustness for machine learning methods. Other
methods for generating these natural perturbations are likely to provide
additional insights into model robustness. As an example, photo sharing
websites contain a large number of near-duplicate images: pairs
of images of the same scene captured at different times, viewpoints, or from a
different camera~\cite{recht2019imagenet}. More generally, devising
similar, domain-specific strategies to collect, verify, and measure robustness
to natural perturbations in domains such as natural language processing or
speech recognition is a promising direction for future work.




%% file: acknowledgements.tex
We thank Rohan Taori for providing models trained for robustness to image corruptions, and Pavel Tokmakov for his help with training detection models on \datasetorig. This research was generously supported in part by ONR awards N00014-17-1-2191, N00014-17-1-2401, and N00014-18-1-2833, the DARPA Assured Autonomy (FA8750-18-C-0101) and Lagrange (W911NF-16-1-0552) programs, an Amazon AWS AI Research Award, and a gift from Microsoft Research.

%% file: appendix.tex
\section{Source Dataset Overview}
\label{sec:sourcedatasets}
\subsection{\datasetorig{}}
\label{sec:dataset}
The 2015 \datasetorig{} dataset is widely used for training video object detectors \cite{han2016seq} as well as trackers \cite{bertinetto2016fully}.
We chose to work with the 2017 \datasetorig{} dataset because it is a superset of the 2015 dataset. 
In total, the 2017 ImageNet-Vid dataset consists of 1,181,113 training frames from 4,000 videos and 512,360 validation frames from 1,314 videos. 
The videos have frame rates ranging from 9 to 59 frames per second (fps), with a median fps of 29. 
The videos range from 0.44 to 96 seconds in duration with a median duration of 12 seconds. 
Each frame is annotated with labels indicating the presence or absence of 30 object classes and corresponding bounding boxes for any label present in the frame.
The 30 classes are ancestors of 293 of the 1,000 ILSVRC-2012 classes.

\subsection{\datasetshlensorig}
\label{sec:datasetshlens}
The 2017 \datasetshlensorig{} is a a large scale dataset with 8,146,143 annotated training frames 253,569 unique videos
and with 1,013,246 validation frames from 31,829 videos. The video segments are approximately 19 seconds long on average. Each frame is annotated
with exactly one label indicating the presence of 22 object classes, all of which are ancestors of 229 out of the ILSVRC-2012 classes.

\section{Full Original vs Perturbed Accuracies}
\label{app:table}
\subsection{\dataset}
{
\label{app:full_clf_results_imvid}
\input{table_appendix_imagenet_vid_robust.tex}
}

\subsection{\datasetshlens}
{
\label{app:full_clf_results_ytbb}
\input{table_appendix_ytbb.tex}
}

\section{Model independent distribution shift}
\label{sec:transfer}
\begin{figure}[ht!]
    \centering
    \begin{subfigure}{0.4\textwidth}
    \includegraphics[width=\textwidth]{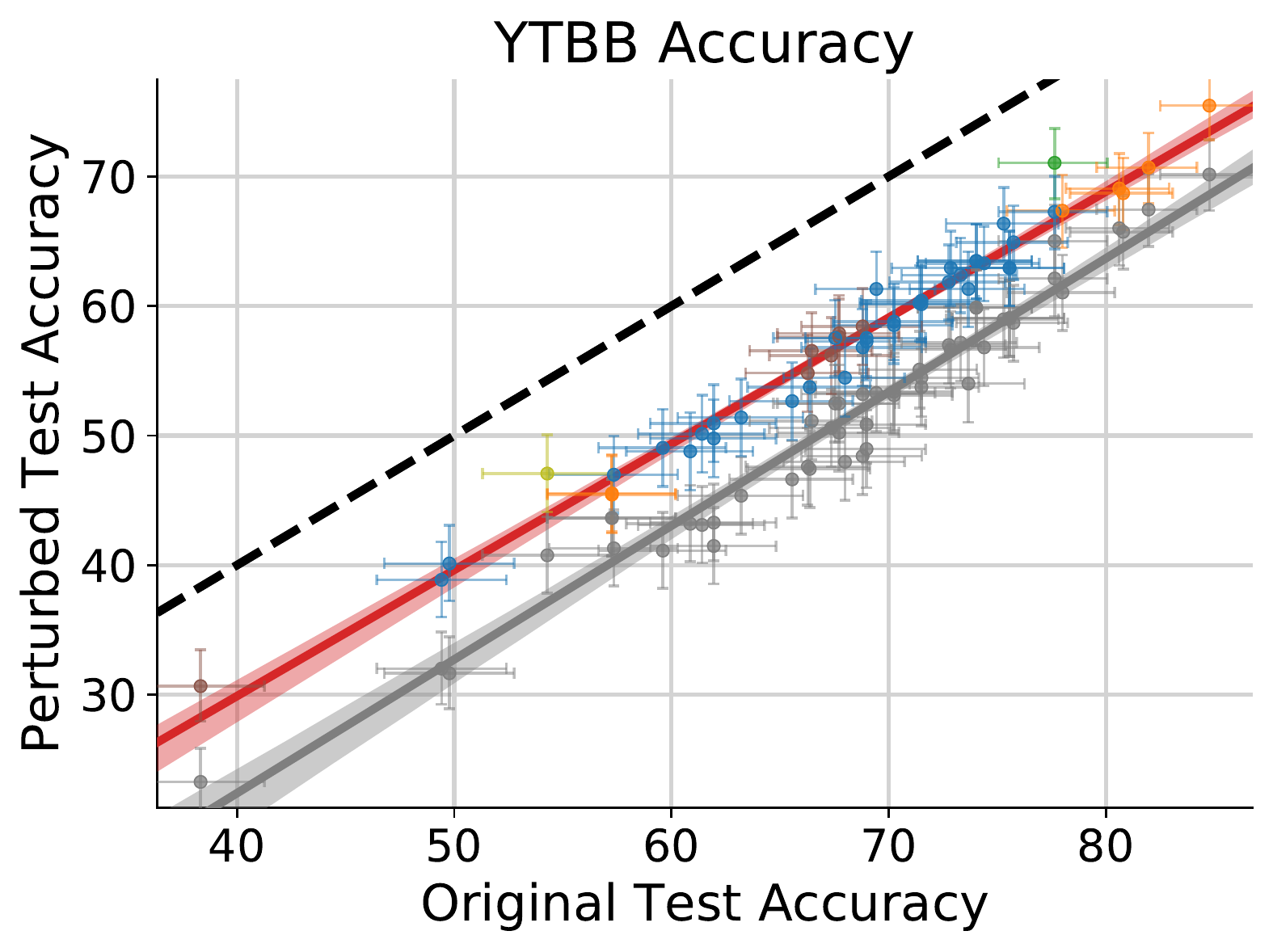}
    \end{subfigure}
    \begin{subfigure}{0.4\textwidth}
    \includegraphics[width=\textwidth]{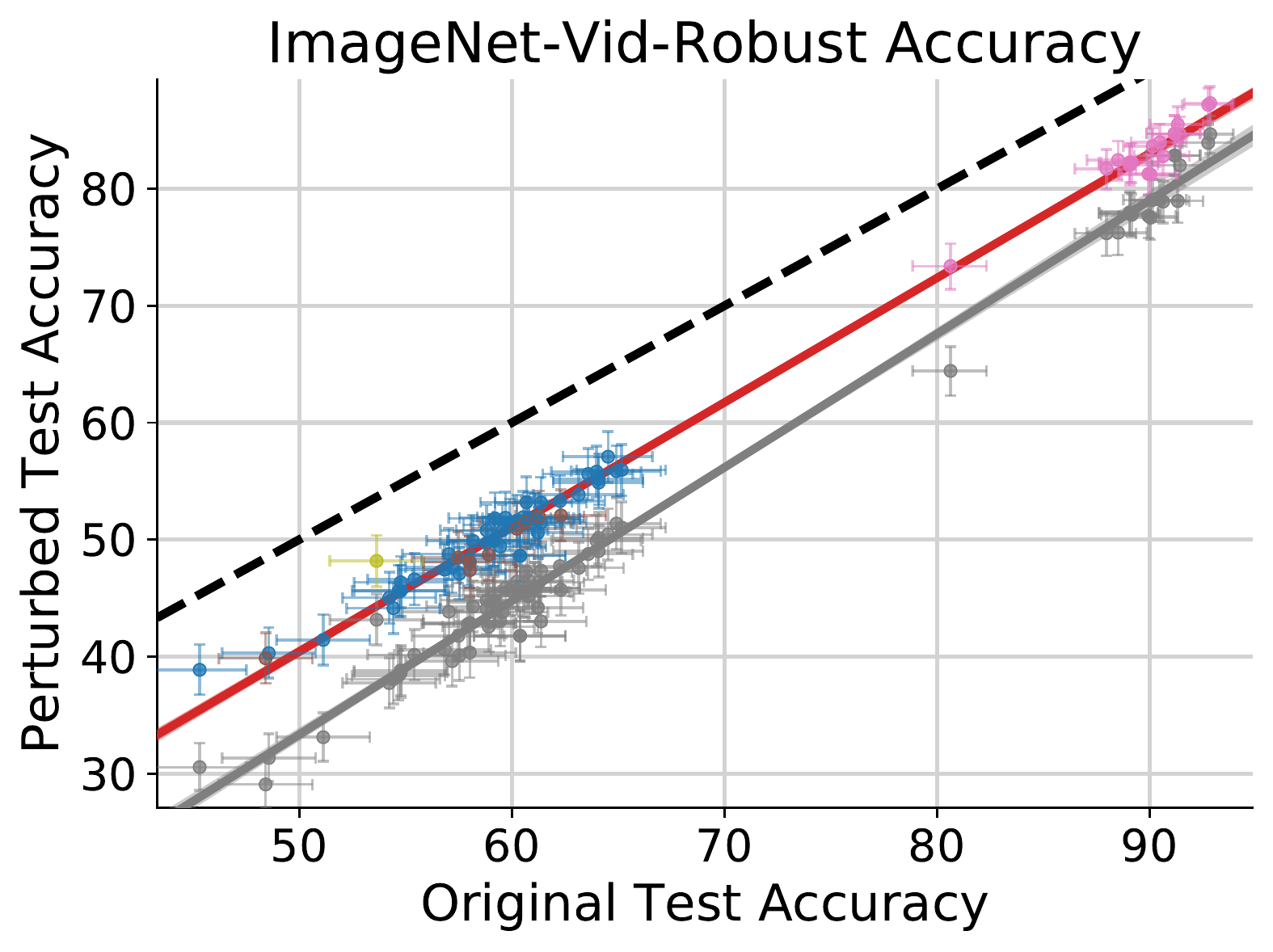}
    \end{subfigure}
    \begin{subfigure}{0.8\textwidth}
    \includegraphics[width=\textwidth]{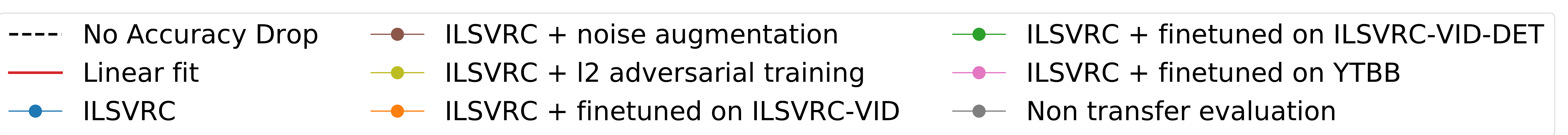}
    \end{subfigure}
    \caption{Model accuracy on original vs. perturbed images for a static
    set of perturbed frames across all models. The grey points and grey linear
    fit correspond to the perturbed accuracies of models evaluated on per model
    perturbations studied in Figure \ref{fig:pm10}
    \label{fig:transfer}}

\end{figure}

Though the distribution shift we induced in our study
were \emph{model dependent} because we found the worst
neighbor frame \emph{for each} model, we could study the
same problem but impose a static set of perturbed frames
across all models. In Figure ~\ref{fig:transfer} we study
this static set of perturbations across all models and see 
a substantial (but smaller) drop in accuracy for both models.
The static set of perturbations were chosen by choosing the 
neighbor frame that the largest number of models got incorrectly.

\section{Per class analysis}
\begin{figure}
    \centering
    \includegraphics[height=0.2\textheight]{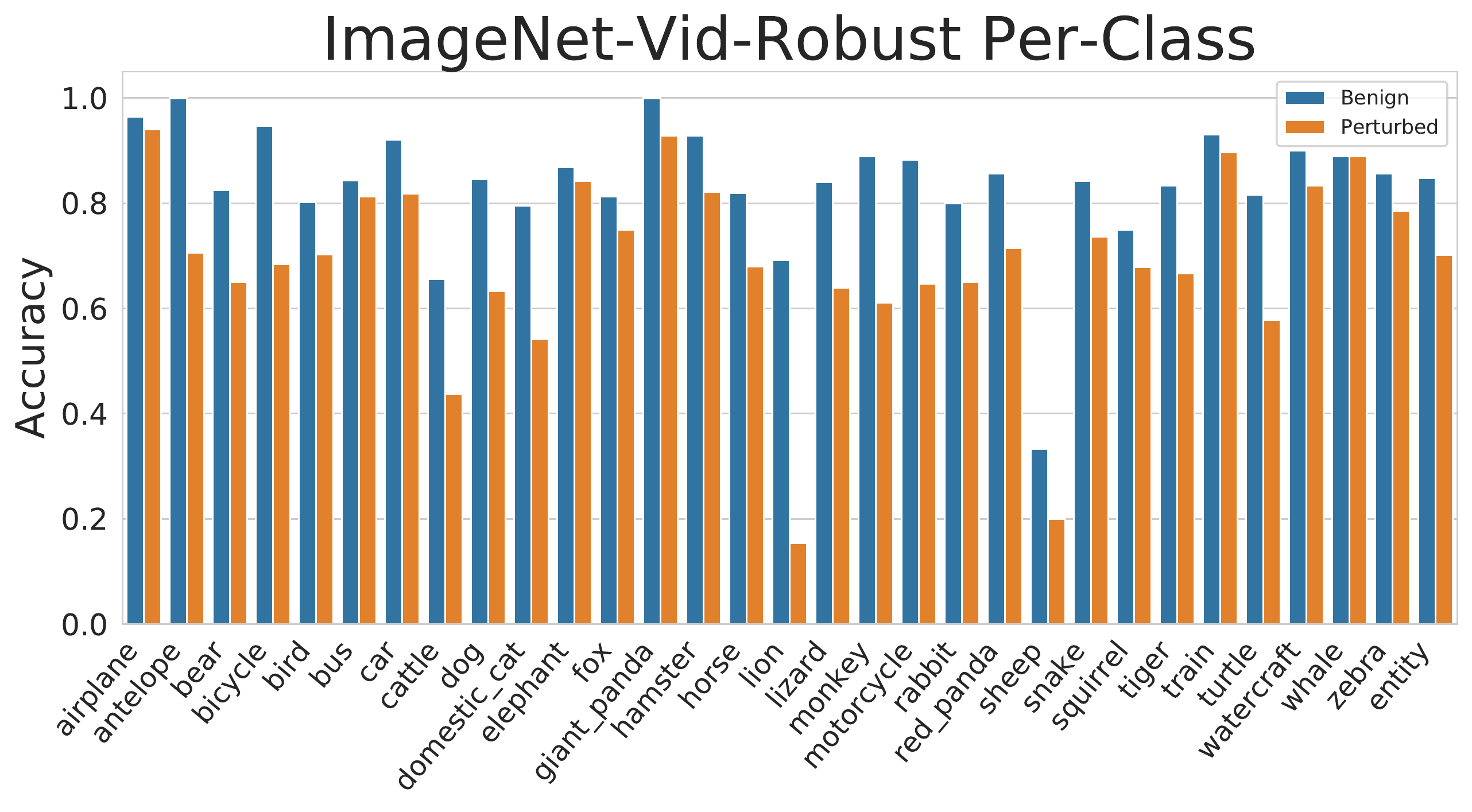}
    \includegraphics[height=0.2\textheight]{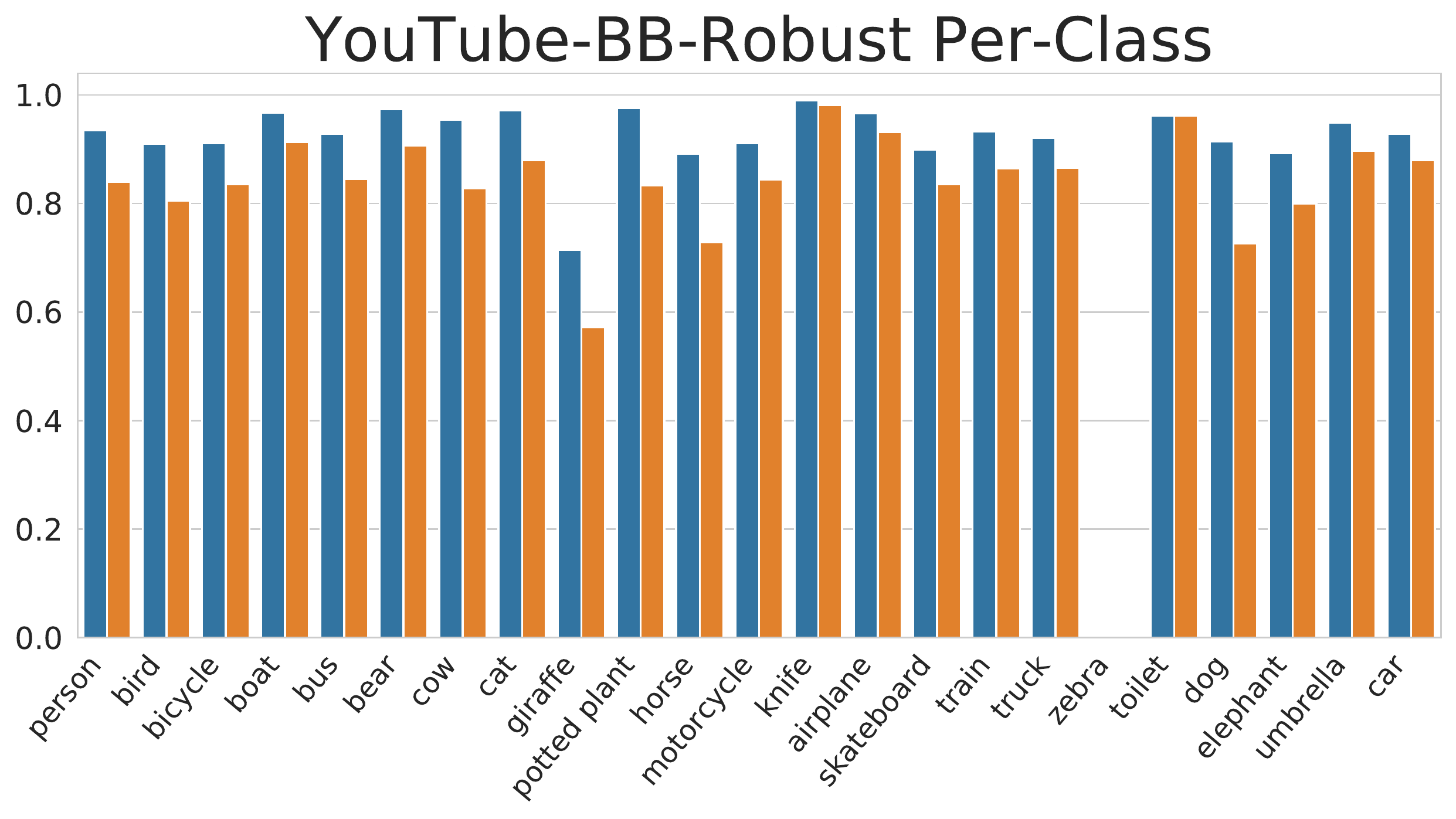}
    \caption{Per-class accuracy statistics for our best performing classification model (fine-tuned ResNet152) on {\dataset} and \datasetshlens{}. For \datasetshlensorig{}, note that `zebra' is the least common label, present in only 24 anchor frames sampled by \cite{gu2019using}, of which 4 are included in our dataset.}
    \label{fig:perclass}
\end{figure}

We study the effect of our perturbations on the 30 classes in \dataset{} and \datasetshlens{} to determine whether the performance drop was concentrated in a few ``hard'' classes.
Figure \ref{fig:perclass} shows the original and perturbed accuracies across classes for our best performing model (a fine-tuned ResNet-152).
Although there are a few particularly difficult classes for perturbed accuracy (e.g., lion or monkey on \dataset{}), the accuracy drop is spread across most classes.
On \dataset{}, this model saw a total drop of $14.4\%$ between original and perturbed images and a median drop of $14.0\%$ in per-class accuracy. On \datasetshlens{}, the total drop was $8.9\%$ and the median drop was $6.7\%$.

\section{Per-frame conditional robustness metric introduced in \cite{gu2019using}}
\label{sec:shlens_compare}
\begin{figure}[ht!]
    \centering
    \includegraphics[height=0.2\textheight]{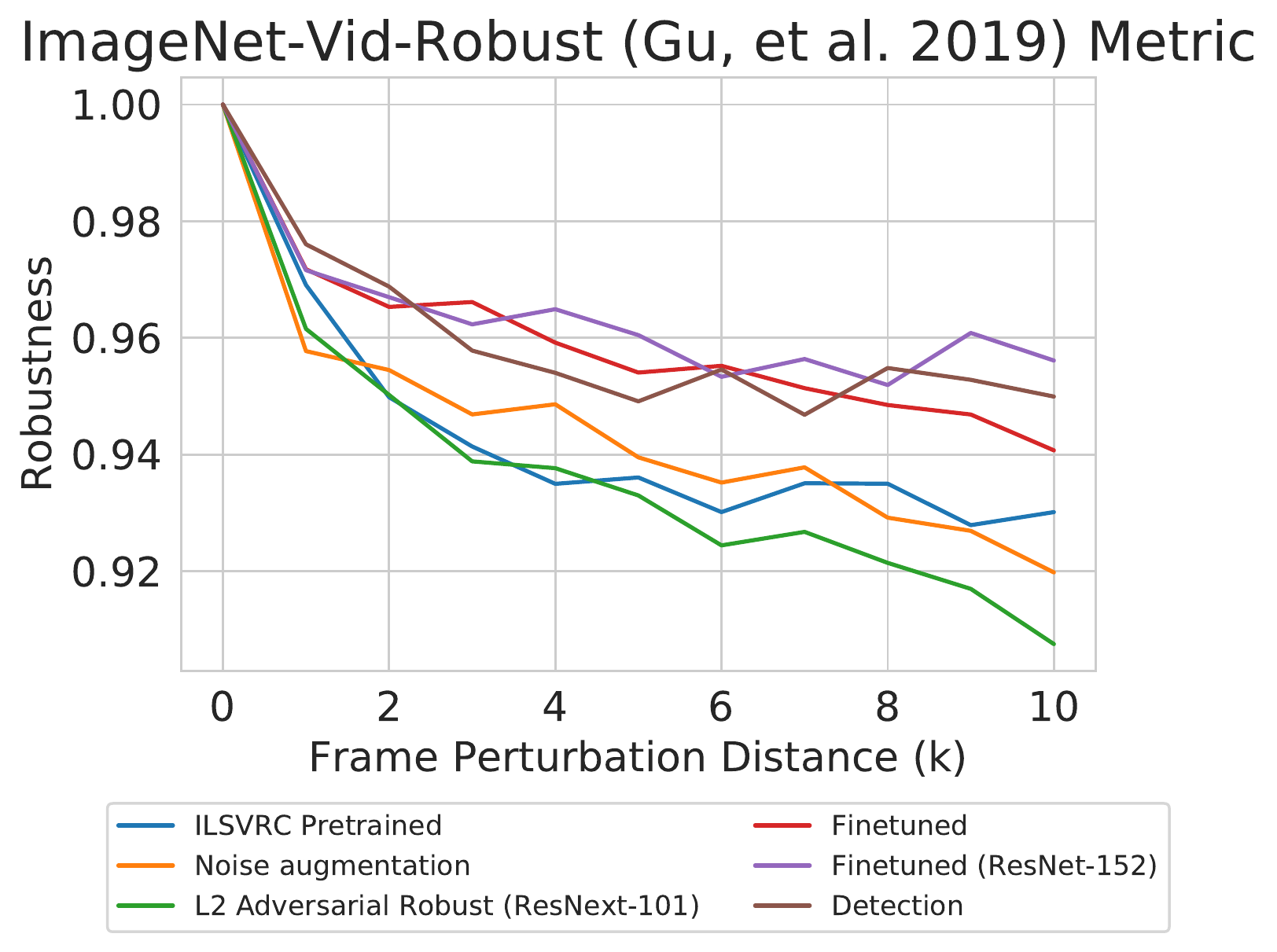}
    \includegraphics[height=0.2\textheight]{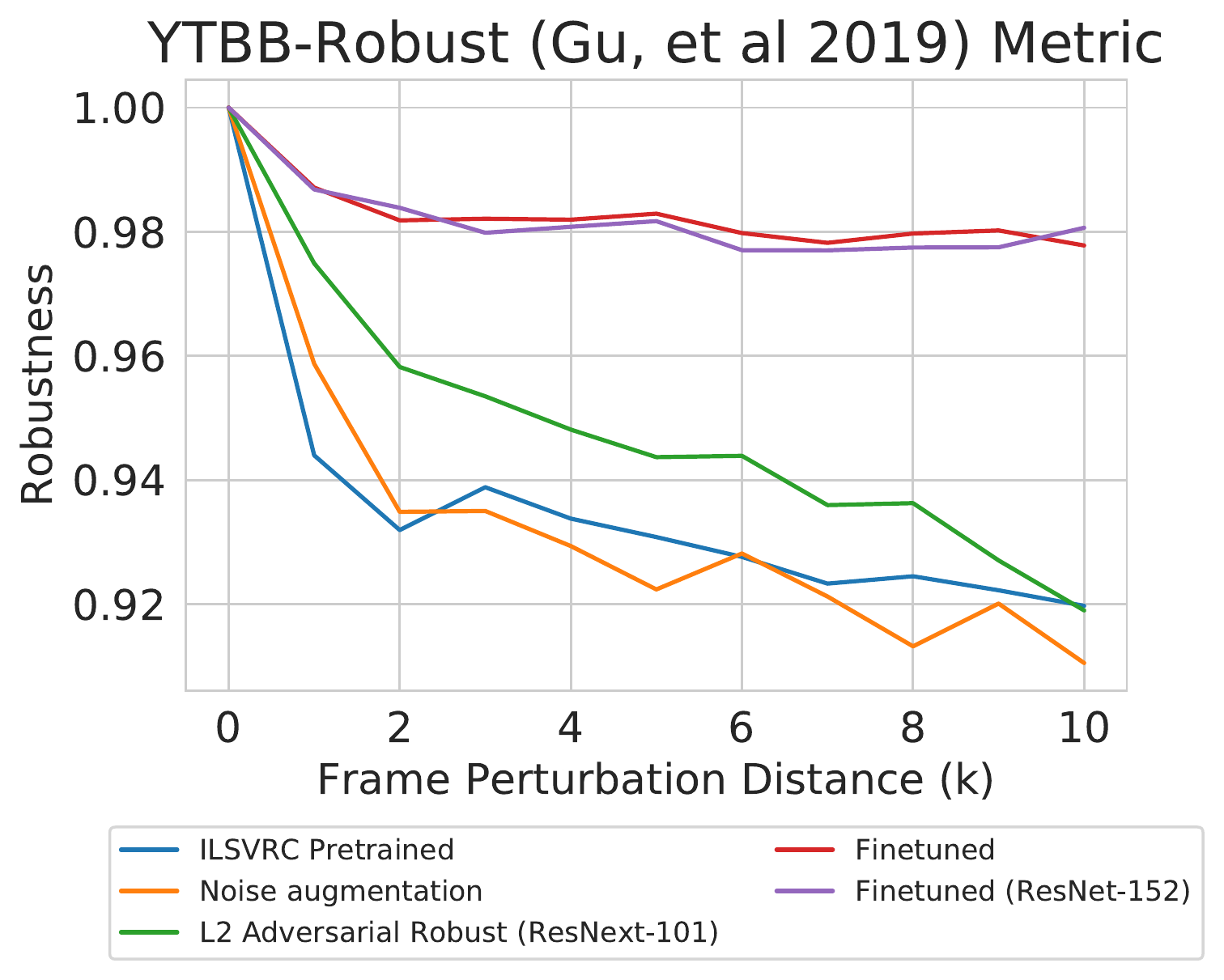}
    \caption{Conditional robustness metric from \cite{gu2019using} on perturbed frames as a function of perturbation distance on \dataset{} and \datasetshlens{}. Model accuracies from five different model types and the best performing model are shown. The model architecture is ResNet-50 unless otherwise mentioned. 
        \label{fig:shlensk}}
\end{figure}

\begin{figure}[ht!]
    \centering
    \includegraphics[height=0.2\textheight]{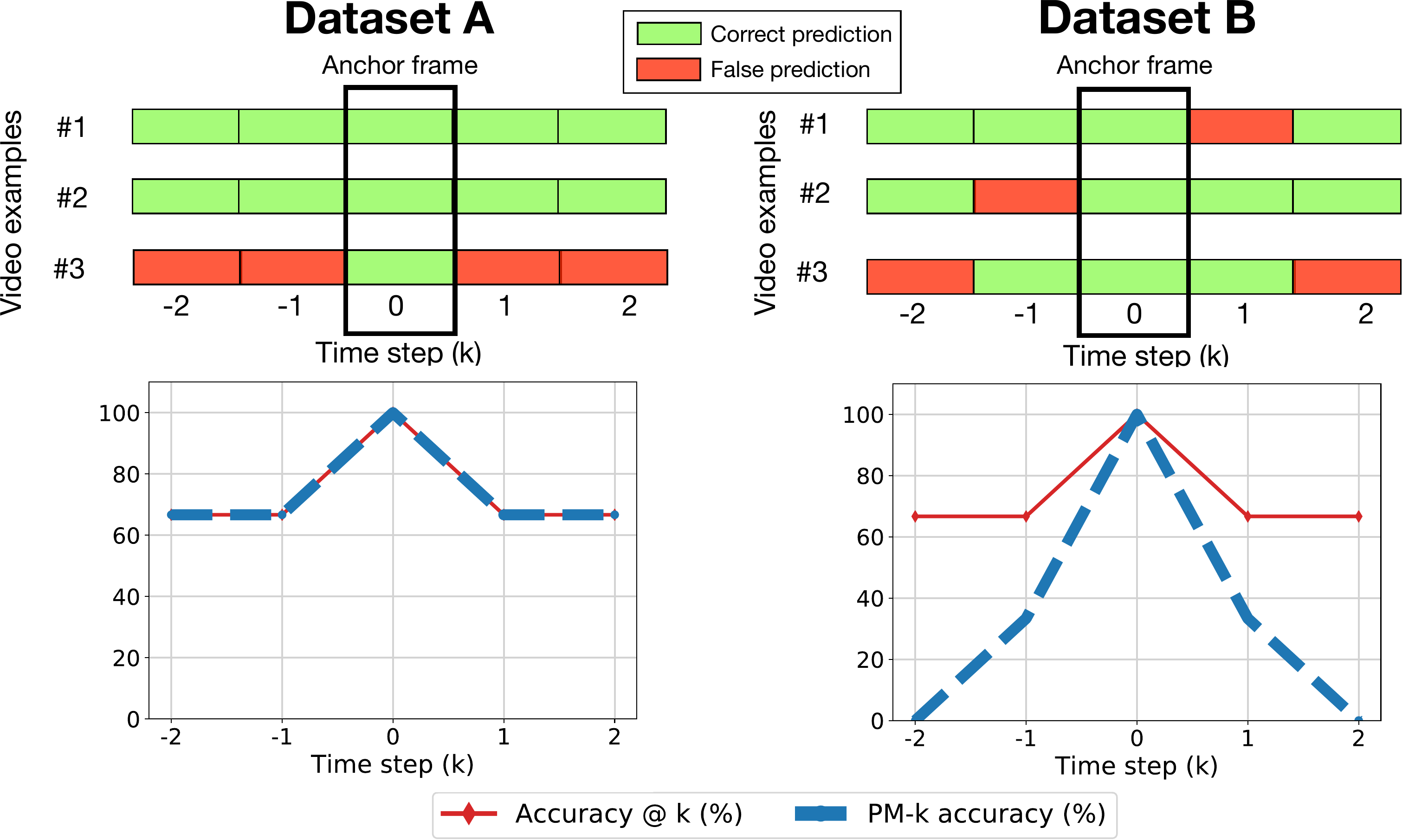}
    \caption{\label{fig:shlens_explain} For the two example videos above the score from \cite{gu2019using} metric (Accuracy @ K) is identical, 
        but the {\pmk} metric behaves substantially differently when the errors are spread across many independent videos, as shown in the right example}
\end{figure}

In concurrent work, the authors of \cite{gu2019using} considered a different metric of robustness. In this section, we compute this metric on all models in our test bed to compare our findings to \cite{gu2019using}.
There are two main differences between \pmk~and the robustness metric in \cite{gu2019using}.
\begin{enumerate}
    \item For two visually similar ``neighbor'' frames $I_{0}$ and $I_{1}$ with true label $Y$ and classifier $f$,  \cite{gu2019using} studies the conditional probability $P(f(I_{1}) = y | f(I_{0})  = y)$
    \item While {\pmk} looks for errors in all neighbor frames in a neighborhood of $k$ frames away from the anchor frame (so this would include frames 1, 2, \dots, k frames away), \cite{gu2019using} only considers errors from \textbf{exactly} k frames away.
\end{enumerate}
In Fig.~\ref{fig:shlens_explain} we illustrate simple example where two videos can have the same behavior for the metric introduced
by \cite{gu2019using} but drastically different behavior for the \pmk metric.

\section{$\ell_{\infty}$ distance vs {\pmk} Accuracy}
\begin{figure}[ht!]
    \includegraphics[width=\textwidth]{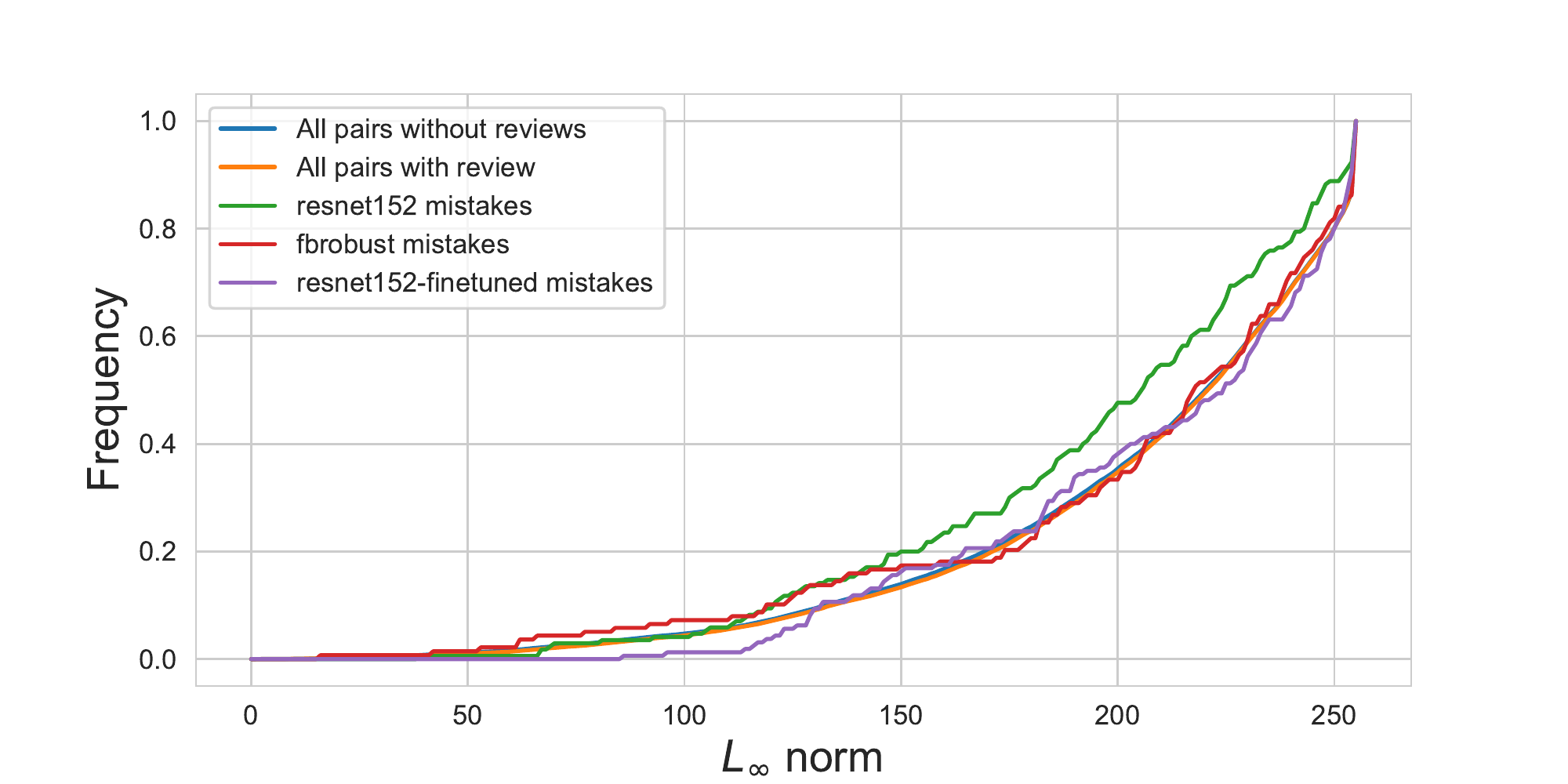}
    \caption{CDF showing the $\ell_{\infty}$ distance between pairs of frames from different distributions.}
    \label{fig:linf_vs_frequency}
\end{figure}

$\ell_{\infty}$ adversarial examples are well studied in the robustness community, yet the connection between
$\ell_{\infty}$ and other forms of more ``natural'' robustness is unclear. Here, we plot the cumulative distribution of the $\ell_{\infty}$ distance
between pairs of nearby frames in our datasets. In~\Cref{fig:linf_vs_frequency}, we show the CDF of $\ell_{\infty}$ distance for all pairs, all reviewed pairs, and mistakes made by 3
indicative models. Note the \verb|fbrobust| model is trained specifically to be robust to $\ell_{\infty}$ adversaries.

\section{{\pmk} Accuracy with varying k}
\subsection{\dataset}
    \begin{figure}[ht!]
    \centering
    \begin{subfigure}{0.45\textwidth}
    \includegraphics[width=\textwidth]{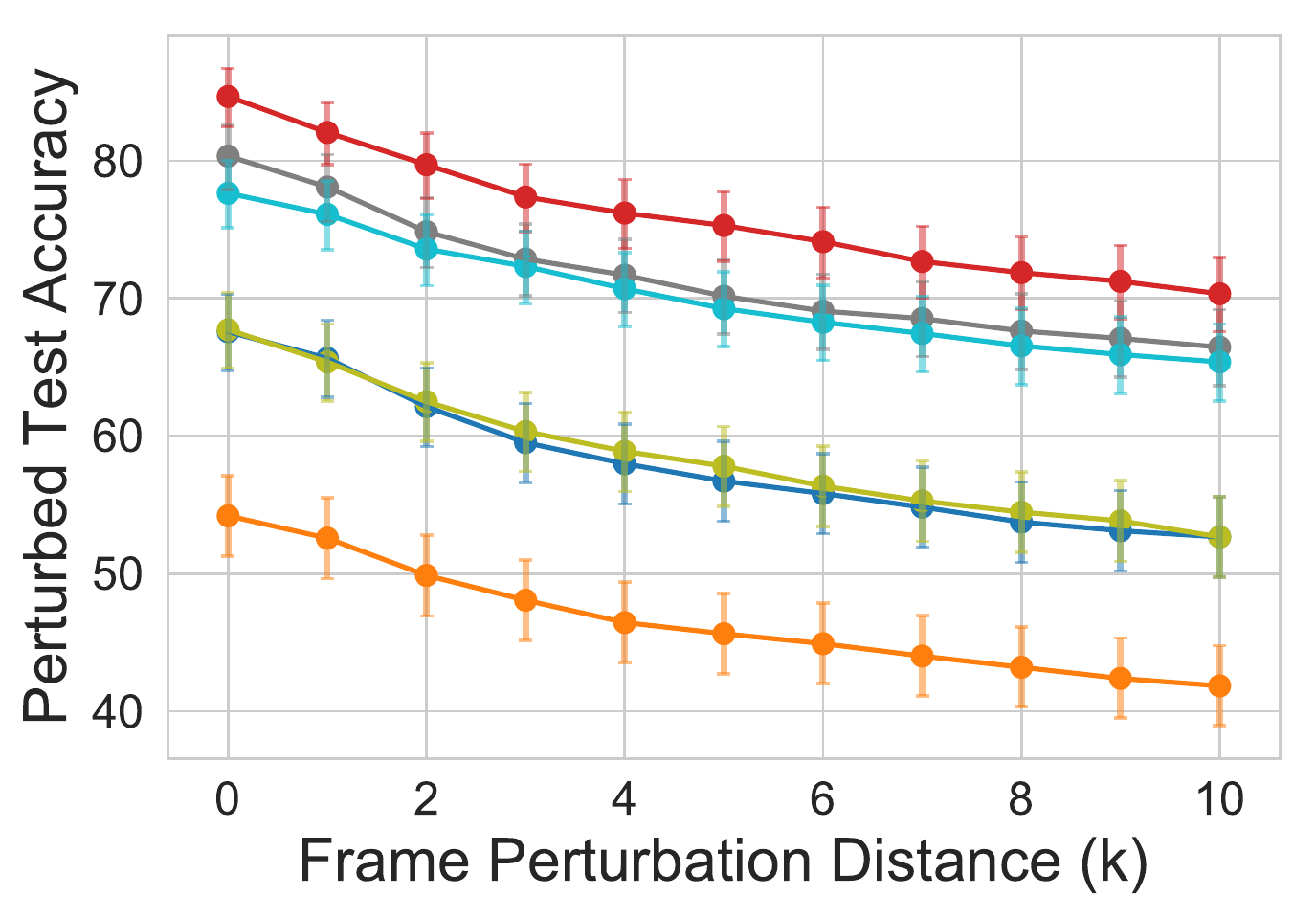}
    \end{subfigure}
    \begin{subfigure}{0.54\textwidth}
    \includegraphics[width=\textwidth]{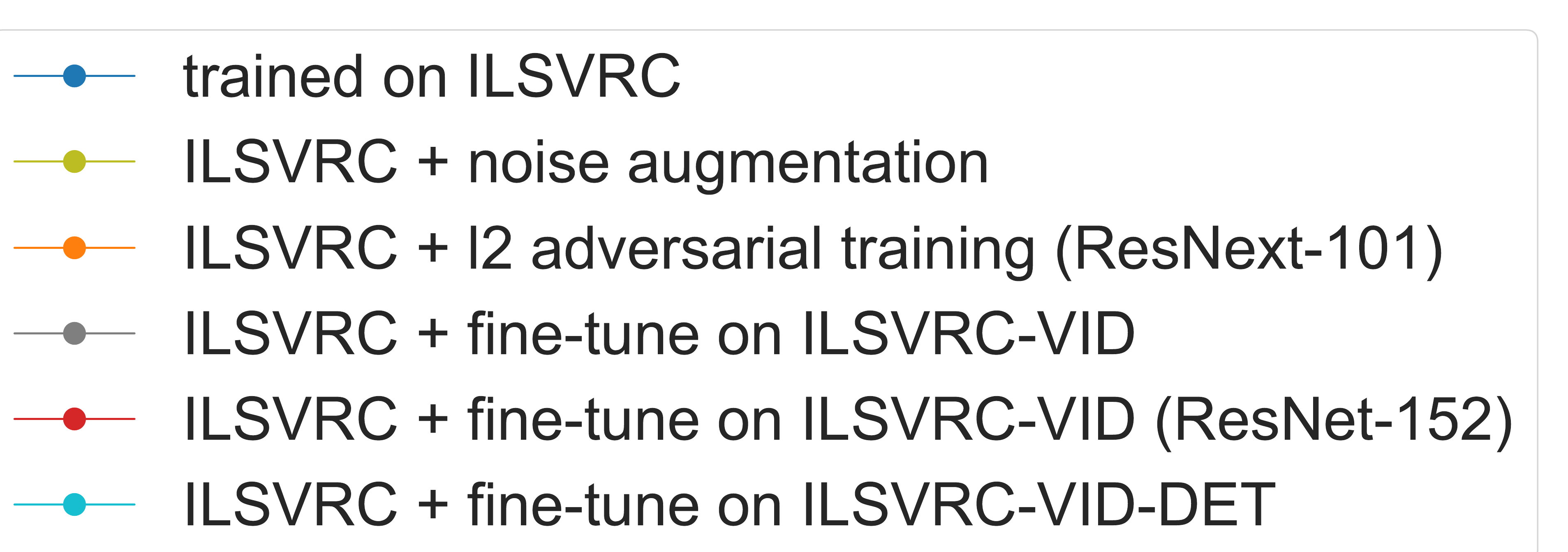}
    \end{subfigure}
    \caption{Model classification accuracy on perturbed frames as a function of perturbation distance (shown with 95\% Clopper-Pearson confidence intervals). Model accuracies from five different model types and the best performing model are shown. The model architecture is ResNet-50 unless otherwise mentioned. 
      \label{fig:pmk}}
    \end{figure}
In \Cref{fig:pmk}, we plot the relationship between {\acca} and perturbation distance (i.e., the  \texttt{k} in the \texttt{pm-k} metric). 
The entire x-axis in \Cref{fig:pmk} corresponds to a temporal distance of at most $0.3$ seconds between the original and perturbed frames.


\section{I-frames and P-Frames}
\subsection{\dataset}
\label{sec:iframe-imagenetvid}
One possible concern with analyzing performance on video frames is the impact of video compression on model robustness. In particular, the videos in \dataset{} contain 3 different frame types: `i-frames', `p-frames', and `b-frames'. `p-frames' are compressed by referencing pixel content from previous frames, while `b-frames' are compressed via references to previous and future frames. `i-frames' are stored without references to other frames.

We compute the original and perturbed accuracies, and the drop in accuracy for a subset of the dataset without `i-frames', a subset without `p-frames', and a subset without `b-frames' in \Cref{tab:iframe-imagenetvid}. While there are modest differences in accuracy due to compression, this analysis suggests that the sensitivity of models is not significantly due to the differences in quality of frames due to video compression.

\begin{table}
    \caption{Analyzing results based on frame-type in video compression. See \Cref{sec:iframe-imagenetvid} for details.}
    \label{tab:iframe-imagenetvid}
    \centering
    \begin{tabular}{lcccccc}
        \toprule
                    & Original Acc. & Perturbed Acc. & $\Delta$ & \# anchor frames \\\midrule
    All frames      & 84.8          & 70.2           & 14.6     & 1109 \\
    w/o `i-frames'  & 84.7          & 70.3           & 14.4     & 1104 \\
    w/o `p-frames'  & 83.9          & 73.7           & 10.2     &  415 \\
    w/o `b-frames'  & 85.4          & 73.2           & 12.2     &  699 \\\bottomrule
    \end{tabular}
\end{table}

\section{FPS analysis}
\subsection{\dataset} To analyze the impact of frame-rate on accuracy, we show results on subsets of videos with fixed fps (25, 29, and 30, which cover 89\% of the dataset) using a fine-tuned ResNet-152 model in \Cref{tab:fps-imagenetvid}. The accuracy drop is similar across the subsets, and similar to the drop for the whole dataset.

\begin{table}[h!]
    \centering
    \begin{tabular}{lcccc}
        \toprule
FPS & Acc. Orig. & Acc. Perturbed  & Drop & \# Videos \\
\midrule
25 & 87.3 \textcolor{gray}{[83.0, 90.9]} & 73.3 \textcolor{gray}{[67.8, 78.3]} & 14.0 & 292 \\
29 & 87.7 \textcolor{gray}{[84.0, 90.8]} & 74.9 \textcolor{gray}{[70.3, 79.2]} & 12.8 & 383 \\
30 & 78.3 \textcolor{gray}{[73.3, 82.7]} & 61.7 \textcolor{gray}{[56.0, 67.1]} & 16.6 & 313 \\
\bottomrule
    \end{tabular}
    \caption{Results on subsets of ImageNet-Vid-Robust with fixed FPS.}
    \label{tab:fps-imagenetvid}
\end{table}


\section{ILSVRC training with \dataset{} classes}
We trained ResNet-50 from scratch on ILSVRC using the 30 ImageNet-Vid classes. We also fine-tuned the model on ImageNet-Vid. In Table \ref{tab:ilsvrc-30}, we show the accuracy drops are consistent with models in our submission. We hypothesize that the lower accuracy is due to coarser supervision on ILSVRC.

\begin{table}[h!]
    \centering
    \begin{tabular}{lccc}
        \toprule
        Model & Acc. Orig. & Acc. Perturbed & Drop \\\midrule
        ILSVRC-30 & 61.0 & 44.9 & 15.1 \\
        ILSVRC-30 + FT & 77.8 & 59.9 & 17.9 \\
        \bottomrule
    \end{tabular}
    \caption{Results of training ResNet-50 on ILSVRC with 30 classes from \dataset.}
    \label{tab:ilsvrc-30}
\end{table}

\section{Experimental Details \& Hyperparameters}
\label{app:expdetails}
All classification experiments were carried out using PyTorch version 1.0.1
on an AWS p3.2xlarge with the NVIDIA V100 GPU. All pretrained models were
downloaded from \cite{cadene} at commit hash \texttt{021d97897c9aa76ec759deff43d341c4fd45d7ba}. Evaluations in
Tables ~\ref{app:full_clf_results_imvid} and \ref{app:full_clf_results_ytbb} all use the default settings for
evaluation. The hyperparameters for the \textit{fine-tuned} models are
presented in Table ~\ref{apdx:hyperparams}. We searched for learning rates
between $10^{-3}$ and $10^{-5}$ for all models.

We additionally detail hyperparameters for detection models in
\Cref{apdx:detection-hyperparams}. Detection experiments were conducted with
PyTorch version 1.0.1 on a machine with 4 Titan X GPUs, using the
Mask R-CNN benchmark
repository\cite{massa2018mrcnn}. We used the default learning rate provided
in \cite{massa2018mrcnn}. For R-FCN, we used the model trained by
\cite{xiao2018video}.

\begin{table}[h!]
\centering
\rowcolors{2}{}{gray!25}
\caption{Hyperparameters for models finetuned on ImageNet-Vid, \label{apdx:hyperparams}}
\begin{tabular}{lccccc}
    \toprule
    Model & Base Learning Rate & Learning Rate Schedule & \thead{Batch Size} &   \thead{Epochs} \\
    \midrule
    resnet152 & $10^{-4}$ & \texttt{Reduce LR On Plateau} & 32 & 10 \\
    resnet50 & $10^{-4}$ & \texttt{Reduce LR On Plateau} & 32 & 10 \\ 
    alexnet & $10^{-5}$ & \texttt{Reduce LR On Plateau} & 32 & 10 \\
    vgg16 & $10^{-5}$ & \texttt{Reduce LR On Plateau} & 32 & 10 \\
    \bottomrule
\end{tabular}
\vspace{-0.6cm}
\end{table}

\begin{table}[h!]
\centering
\rowcolors{2}{}{gray!25}
\caption{Hyperparameters for detection models.\label{apdx:detection-hyperparams}}
\begin{tabular}{lccccc}
    \toprule
    \rowcolor{white!50}
    Model & Base Learning Rate & Learning Rate Schedule & \thead{Batch Size} &   \thead{Iterations} \\
    \midrule
    F-RCNN ResNet-50  & $10^{-2}$ & \texttt{Step 20k, 30k} & 8 & 40k \\
    F-RCNN ResNet-101 & $10^{-2}$ & \texttt{Step 20k, 30k} & 8 & 40k \\ 
    \bottomrule
\end{tabular}
\vspace{-0.6cm}
\end{table}

\section{Detection pm-k}
We briefly introduce the mAP metric for detection here and refer the reader to~\cite{cocomapsite} for further details.
The standard detection metric proceeds by first determining whether each predicted bounding box in an image is a true or false positive, based on the intersection over union (IoU) of the predicted and ground truth bounding boxes.
The metric then computes the per-category average precision (AP, averaged over recall thresholds) of the predictions across all images.
The final metric is reported as the mean of these per-category APs (mAP).

We define the \texttt{pm-k} analog of mAP by replacing each anchor frame in the dataset with a nearby frame that minimizes the per-image average precision. Since the category-specific average precision is undefined for categories not present in an image, we minimize the average precision across categories present in each frame rather than the mAP.

%% file: table_appendix_imagenet_vid_robust.tex
	\rowcolors{3}{}{gray!25}
	\begin{longtable}{lcccc}
	    \toprule
	Model & \thead{Accuracy \\ Original} & \thead{Accuracy \\ Perturbed}  & $\Delta$ \\
	\midrule
    resnet152\_finetuned & 84.8 {\footnotesize \textcolor{gray}{[82.5, 86.8]}} & 70.2 {\footnotesize \textcolor{gray}{[67.4, 72.8]}} & 14.6\\
resnet50\_finetuned & 80.8 {\footnotesize \textcolor{gray}{[78.3, 83.1]}} & 65.7 {\footnotesize \textcolor{gray}{[62.9, 68.5]}} & 15.1\\
vgg16bn\_finetuned & 78.0 {\footnotesize \textcolor{gray}{[75.4, 80.4]}} & 61.0 {\footnotesize \textcolor{gray}{[58.1, 63.9]}} & 17.0\\
nasnetalarge\_imagenet\_pretrained & 77.6 {\footnotesize \textcolor{gray}{[75.1, 80.1]}} & 62.1 {\footnotesize \textcolor{gray}{[59.2, 65.0]}} & 15.5\\
resnet50\_detection & 77.6 {\footnotesize \textcolor{gray}{[75.1, 80.1]}} & 65.0 {\footnotesize \textcolor{gray}{[62.1, 67.8]}} & 12.6\\
inceptionresnetv2\_imagenet\_pretrained & 75.7 {\footnotesize \textcolor{gray}{[73.1, 78.2]}} & 58.7 {\footnotesize \textcolor{gray}{[55.7, 61.6]}} & 17.0\\
dpn107\_imagenet\_pretrained & 75.6 {\footnotesize \textcolor{gray}{[72.9, 78.1]}} & 59.1 {\footnotesize \textcolor{gray}{[56.1, 62.0]}} & 16.5\\
inceptionv4\_imagenet\_pretrained & 75.3 {\footnotesize \textcolor{gray}{[72.6, 77.8]}} & 59.0 {\footnotesize \textcolor{gray}{[56.0, 61.9]}} & 16.3\\
dpn92\_imagenet\_pretrained & 74.4 {\footnotesize \textcolor{gray}{[71.7, 76.9]}} & 56.8 {\footnotesize \textcolor{gray}{[53.8, 59.7]}} & 17.6\\
dpn131\_imagenet\_pretrained & 74.0 {\footnotesize \textcolor{gray}{[71.3, 76.6]}} & 59.9 {\footnotesize \textcolor{gray}{[56.9, 62.8]}} & 14.1\\
dpn68b\_imagenet\_pretrained & 73.7 {\footnotesize \textcolor{gray}{[71.0, 76.2]}} & 54.0 {\footnotesize \textcolor{gray}{[51.0, 57.0]}} & 19.7\\
resnext101\_32x4d\_imagenet\_pretrained & 73.3 {\footnotesize \textcolor{gray}{[70.6, 75.9]}} & 57.2 {\footnotesize \textcolor{gray}{[54.2, 60.1]}} & 16.1\\
resnext101\_64x4d\_imagenet\_pretrained & 72.9 {\footnotesize \textcolor{gray}{[70.1, 75.5]}} & 56.6 {\footnotesize \textcolor{gray}{[53.7, 59.6]}} & 16.3\\
resnet152\_imagenet\_pretrained & 72.8 {\footnotesize \textcolor{gray}{[70.0, 75.4]}} & 57.0 {\footnotesize \textcolor{gray}{[54.0, 59.9]}} & 15.8\\
resnet101\_imagenet\_pretrained & 71.5 {\footnotesize \textcolor{gray}{[68.7, 74.1]}} & 53.7 {\footnotesize \textcolor{gray}{[50.8, 56.7]}} & 17.8\\
fbresnet152\_imagenet\_pretrained & 71.5 {\footnotesize \textcolor{gray}{[68.7, 74.1]}} & 54.5 {\footnotesize \textcolor{gray}{[51.5, 57.4]}} & 17.0\\
densenet161\_imagenet\_pretrained & 71.4 {\footnotesize \textcolor{gray}{[68.7, 74.1]}} & 55.1 {\footnotesize \textcolor{gray}{[52.1, 58.1]}} & 16.3\\
densenet169\_imagenet\_pretrained & 70.2 {\footnotesize \textcolor{gray}{[67.5, 72.9]}} & 53.1 {\footnotesize \textcolor{gray}{[50.1, 56.1]}} & 17.1\\
densenet201\_imagenet\_pretrained & 70.2 {\footnotesize \textcolor{gray}{[67.5, 72.9]}} & 53.4 {\footnotesize \textcolor{gray}{[50.4, 56.4]}} & 16.8\\
dpn68\_imagenet\_pretrained & 69.4 {\footnotesize \textcolor{gray}{[66.6, 72.1]}} & 53.3 {\footnotesize \textcolor{gray}{[50.3, 56.3]}} & 16.1\\
bninception\_imagenet\_pretrained & 69.0 {\footnotesize \textcolor{gray}{[66.2, 71.7]}} & 49.0 {\footnotesize \textcolor{gray}{[46.0, 51.9]}} & 20.0\\
densenet121\_imagenet\_pretrained & 69.0 {\footnotesize \textcolor{gray}{[66.2, 71.7]}} & 50.9 {\footnotesize \textcolor{gray}{[47.9, 53.8]}} & 18.1\\
nasnetamobile\_imagenet\_pretrained & 68.8 {\footnotesize \textcolor{gray}{[66.0, 71.5]}} & 48.4 {\footnotesize \textcolor{gray}{[45.4, 51.4]}} & 20.4\\
resnet50\_augment\_\_\_jpeg\_compression & 68.8 {\footnotesize \textcolor{gray}{[66.0, 71.5]}} & 53.2 {\footnotesize \textcolor{gray}{[50.2, 56.2]}} & 15.6\\
resnet34\_imagenet\_pretrained & 68.0 {\footnotesize \textcolor{gray}{[65.2, 70.7]}} & 48.0 {\footnotesize \textcolor{gray}{[45.0, 51.0]}} & 20.0\\
resnet50\_augment\_\_\_impulse\_noise & 67.7 {\footnotesize \textcolor{gray}{[64.9, 70.5]}} & 50.2 {\footnotesize \textcolor{gray}{[47.2, 53.2]}} & 17.5\\
resnet50\_augment\_\_gaussian\_blur & 67.7 {\footnotesize \textcolor{gray}{[64.9, 70.5]}} & 52.5 {\footnotesize \textcolor{gray}{[49.5, 55.5]}} & 15.2\\
resnet50\_imagenet\_pretrained & 67.5 {\footnotesize \textcolor{gray}{[64.7, 70.3]}} & 52.5 {\footnotesize \textcolor{gray}{[49.5, 55.5]}} & 15.0\\
resnet50\_augment\_\_\_gaussian\_noise & 67.4 {\footnotesize \textcolor{gray}{[64.5, 70.1]}} & 50.6 {\footnotesize \textcolor{gray}{[47.6, 53.6]}} & 16.8\\
resnet50\_augment\_\_\_shot\_noise & 66.5 {\footnotesize \textcolor{gray}{[63.6, 69.2]}} & 51.1 {\footnotesize \textcolor{gray}{[48.1, 54.1]}} & 15.4\\
vgg16\_bn\_imagenet\_pretrained & 66.4 {\footnotesize \textcolor{gray}{[63.5, 69.1]}} & 47.4 {\footnotesize \textcolor{gray}{[44.5, 50.4]}} & 19.0\\
resnet50\_augment\_\_\_defocus\_blur & 66.3 {\footnotesize \textcolor{gray}{[63.4, 69.1]}} & 47.6 {\footnotesize \textcolor{gray}{[44.6, 50.6]}} & 18.7\\
vgg19\_bn\_imagenet\_pretrained & 65.6 {\footnotesize \textcolor{gray}{[62.7, 68.4]}} & 46.6 {\footnotesize \textcolor{gray}{[43.6, 49.6]}} & 19.0\\
vgg19\_imagenet\_pretrained & 63.2 {\footnotesize \textcolor{gray}{[60.3, 66.1]}} & 45.4 {\footnotesize \textcolor{gray}{[42.4, 48.3]}} & 17.8\\
resnet18\_imagenet\_pretrained & 61.9 {\footnotesize \textcolor{gray}{[59.0, 64.8]}} & 41.5 {\footnotesize \textcolor{gray}{[38.6, 44.4]}} & 20.4\\
vgg13\_bn\_imagenet\_pretrained & 61.9 {\footnotesize \textcolor{gray}{[59.0, 64.8]}} & 43.3 {\footnotesize \textcolor{gray}{[40.3, 46.3]}} & 18.6\\
vgg16\_imagenet\_pretrained & 61.4 {\footnotesize \textcolor{gray}{[58.5, 64.3]}} & 43.1 {\footnotesize \textcolor{gray}{[40.2, 46.1]}} & 18.3\\
vgg11\_bn\_imagenet\_pretrained & 60.9 {\footnotesize \textcolor{gray}{[57.9, 63.8]}} & 43.2 {\footnotesize \textcolor{gray}{[40.3, 46.2]}} & 17.7\\
vgg13\_imagenet\_pretrained & 59.6 {\footnotesize \textcolor{gray}{[56.6, 62.5]}} & 41.1 {\footnotesize \textcolor{gray}{[38.2, 44.1]}} & 18.5\\
vgg11\_imagenet\_pretrained & 57.3 {\footnotesize \textcolor{gray}{[54.4, 60.3]}} & 41.3 {\footnotesize \textcolor{gray}{[38.4, 44.3]}} & 16.0\\
alexnet\_finetuned & 57.3 {\footnotesize \textcolor{gray}{[54.3, 60.2]}} & 43.6 {\footnotesize \textcolor{gray}{[40.7, 46.6]}} & 13.7\\
ResNeXtDenoiseAll-101\_robust\_pgd & 54.3 {\footnotesize \textcolor{gray}{[51.3, 57.2]}} & 40.8 {\footnotesize \textcolor{gray}{[37.8, 43.7]}} & 13.5\\
squeezenet1\_1\_imagenet\_pretrained & 49.8 {\footnotesize \textcolor{gray}{[46.8, 52.8]}} & 31.7 {\footnotesize \textcolor{gray}{[28.9, 34.5]}} & 18.1\\
alexnet\_imagenet\_pretrained & 49.4 {\footnotesize \textcolor{gray}{[46.4, 52.4]}} & 32.0 {\footnotesize \textcolor{gray}{[29.3, 34.8]}} & 17.4\\
resnet50\_augment\_\_\_contrast\_change & 38.3 {\footnotesize \textcolor{gray}{[35.5, 41.3]}} & 23.3 {\footnotesize \textcolor{gray}{[20.8, 25.9]}} & 15.0\\
    \bottomrule 
    \hiderowcolors
    \caption{\label{apdx:full_clf_results_imagenet_vid} Classification model perturbed and original accuracies for all models in our test bed evaluated on the ImageNet-Vid-Robust dataset.}
                                                \end{longtable}
    

%% file: table_appendix_ytbb.tex
	\rowcolors{3}{}{gray!25}
	\begin{longtable}{lcccc}
	    \toprule
	Model & \thead{Accuracy \\ Original} & \thead{Accuracy \\ Perturbed}  & $\Delta$ \\
	\midrule
    resnet152\_finetuned & 92.9 {\footnotesize \textcolor{gray}{[91.2, 94.3]}} & 84.7 {\footnotesize \textcolor{gray}{[82.4, 86.8]}} & 8.2\\
resnet50\_finetuned & 91.4 {\footnotesize \textcolor{gray}{[89.6, 93.0]}} & 82.0 {\footnotesize \textcolor{gray}{[79.6, 84.2]}} & 9.4\\
inceptionresnetv2\_finetuned & 91.3 {\footnotesize \textcolor{gray}{[89.5, 92.9]}} & 79.0 {\footnotesize \textcolor{gray}{[76.4, 81.3]}} & 12.3\\
vgg19\_finetuned & 90.5 {\footnotesize \textcolor{gray}{[88.6, 92.2]}} & 79.1 {\footnotesize \textcolor{gray}{[76.5, 81.4]}} & 11.4\\
vgg16\_finetuned & 89.1 {\footnotesize \textcolor{gray}{[87.1, 90.8]}} & 78.0 {\footnotesize \textcolor{gray}{[75.4, 80.4]}} & 11.1\\
inceptionv4\_finetuned & 88.5 {\footnotesize \textcolor{gray}{[86.5, 90.3]}} & 76.3 {\footnotesize \textcolor{gray}{[73.6, 78.7]}} & 12.2\\
resnet18\_finetuned & 88.0 {\footnotesize \textcolor{gray}{[85.9, 89.8]}} & 76.2 {\footnotesize \textcolor{gray}{[73.6, 78.7]}} & 11.8\\
alexnet\_finetuned & 80.6 {\footnotesize \textcolor{gray}{[78.2, 82.9]}} & 64.4 {\footnotesize \textcolor{gray}{[61.5, 67.3]}} & 16.2\\
pnasnet5large\_imagenet\_pretrained & 65.2 {\footnotesize \textcolor{gray}{[62.3, 68.0]}} & 51.0 {\footnotesize \textcolor{gray}{[48.0, 54.0]}} & 14.2\\
nasnetalarge\_imagenet\_pretrained & 64.9 {\footnotesize \textcolor{gray}{[62.0, 67.7]}} & 51.4 {\footnotesize \textcolor{gray}{[48.4, 54.4]}} & 13.5\\
inceptionresnetv2\_imagenet\_pretrained & 64.5 {\footnotesize \textcolor{gray}{[61.6, 67.4]}} & 50.4 {\footnotesize \textcolor{gray}{[47.5, 53.4]}} & 14.1\\
dpn98\_imagenet\_pretrained & 64.1 {\footnotesize \textcolor{gray}{[61.2, 66.9]}} & 49.0 {\footnotesize \textcolor{gray}{[46.0, 52.0]}} & 15.1\\
dpn107\_imagenet\_pretrained & 64.1 {\footnotesize \textcolor{gray}{[61.2, 66.9]}} & 50.1 {\footnotesize \textcolor{gray}{[47.2, 53.1]}} & 14.0\\
dpn131\_imagenet\_pretrained & 64.0 {\footnotesize \textcolor{gray}{[61.1, 66.8]}} & 49.9 {\footnotesize \textcolor{gray}{[46.9, 52.9]}} & 14.1\\
inceptionv4\_imagenet\_pretrained & 63.6 {\footnotesize \textcolor{gray}{[60.7, 66.4]}} & 48.8 {\footnotesize \textcolor{gray}{[45.8, 51.8]}} & 14.8\\
xception\_imagenet\_pretrained & 63.2 {\footnotesize \textcolor{gray}{[60.2, 66.0]}} & 47.6 {\footnotesize \textcolor{gray}{[44.6, 50.6]}} & 15.6\\
dpn92\_imagenet\_pretrained & 62.3 {\footnotesize \textcolor{gray}{[59.3, 65.1]}} & 47.7 {\footnotesize \textcolor{gray}{[44.8, 50.7]}} & 14.6\\
resnet50\_augment\_\_jpeg\_compressioon & 62.3 {\footnotesize \textcolor{gray}{[59.4, 65.2]}} & 45.7 {\footnotesize \textcolor{gray}{[42.8, 48.7]}} & 16.6\\
polynet\_imagenet\_pretrained & 61.4 {\footnotesize \textcolor{gray}{[58.4, 64.3]}} & 47.3 {\footnotesize \textcolor{gray}{[44.4, 50.3]}} & 14.1\\
nasnetamobile\_imagenet\_pretrained & 61.4 {\footnotesize \textcolor{gray}{[58.4, 64.3]}} & 43.0 {\footnotesize \textcolor{gray}{[40.1, 46.0]}} & 18.4\\
resnet50\_augment\_\_shot\_noise & 61.3 {\footnotesize \textcolor{gray}{[58.3, 64.2]}} & 46.4 {\footnotesize \textcolor{gray}{[43.4, 49.3]}} & 14.9\\
dpn68\_imagenet\_pretrained & 61.2 {\footnotesize \textcolor{gray}{[58.3, 64.1]}} & 44.2 {\footnotesize \textcolor{gray}{[41.2, 47.2]}} & 17.0\\
fbresnet152\_imagenet\_pretrained & 61.1 {\footnotesize \textcolor{gray}{[58.1, 64.0]}} & 45.9 {\footnotesize \textcolor{gray}{[42.9, 48.8]}} & 15.2\\
resnet152\_imagenet\_pretrained & 60.8 {\footnotesize \textcolor{gray}{[57.8, 63.7]}} & 46.5 {\footnotesize \textcolor{gray}{[43.5, 49.5]}} & 14.3\\
resnet101\_imagenet\_pretrained & 60.8 {\footnotesize \textcolor{gray}{[57.8, 63.7]}} & 45.2 {\footnotesize \textcolor{gray}{[42.2, 48.2]}} & 15.6\\
senet154\_imagenet\_pretrained & 60.7 {\footnotesize \textcolor{gray}{[57.7, 63.6]}} & 47.2 {\footnotesize \textcolor{gray}{[44.3, 50.2]}} & 13.5\\
resnet50\_augment\_\_impulse\_noise & 60.6 {\footnotesize \textcolor{gray}{[57.7, 63.5]}} & 45.5 {\footnotesize \textcolor{gray}{[42.6, 48.5]}} & 15.1\\
se\_resnet101\_imagenet\_pretrained & 60.5 {\footnotesize \textcolor{gray}{[57.6, 63.4]}} & 45.6 {\footnotesize \textcolor{gray}{[42.6, 48.6]}} & 14.9\\
bninception\_imagenet\_pretrained & 60.4 {\footnotesize \textcolor{gray}{[57.4, 63.3]}} & 41.8 {\footnotesize \textcolor{gray}{[38.9, 44.7]}} & 18.6\\
densenet161\_imagenet\_pretrained & 60.2 {\footnotesize \textcolor{gray}{[57.3, 63.1]}} & 46.4 {\footnotesize \textcolor{gray}{[43.4, 49.4]}} & 13.8\\
resnet50\_augment\_\_gaussian\_noise & 60.2 {\footnotesize \textcolor{gray}{[57.3, 63.1]}} & 45.7 {\footnotesize \textcolor{gray}{[42.8, 48.7]}} & 14.5\\
se\_resnext50\_32x4d\_imagenet\_pretrained & 59.9 {\footnotesize \textcolor{gray}{[56.9, 62.8]}} & 45.7 {\footnotesize \textcolor{gray}{[42.7, 48.6]}} & 14.2\\
dpn68b\_imagenet\_pretrained & 59.7 {\footnotesize \textcolor{gray}{[56.7, 62.6]}} & 45.9 {\footnotesize \textcolor{gray}{[42.9, 48.8]}} & 13.8\\
inceptionv3\_imagenet\_pretrained & 59.6 {\footnotesize \textcolor{gray}{[56.6, 62.5]}} & 43.8 {\footnotesize \textcolor{gray}{[40.8, 46.8]}} & 15.8\\
densenet121\_imagenet\_pretrained & 59.5 {\footnotesize \textcolor{gray}{[56.5, 62.4]}} & 43.1 {\footnotesize \textcolor{gray}{[40.1, 46.0]}} & 16.4\\
se\_resnext101\_32x4d\_imagenet\_pretrained & 59.2 {\footnotesize \textcolor{gray}{[56.3, 62.1]}} & 45.2 {\footnotesize \textcolor{gray}{[42.3, 48.2]}} & 14.0\\
densenet201\_imagenet\_pretrained & 59.2 {\footnotesize \textcolor{gray}{[56.2, 62.1]}} & 44.8 {\footnotesize \textcolor{gray}{[41.8, 47.8]}} & 14.4\\
densenet169\_imagenet\_pretrained & 59.2 {\footnotesize \textcolor{gray}{[56.2, 62.1]}} & 44.6 {\footnotesize \textcolor{gray}{[41.7, 47.6]}} & 14.6\\
resnet50\_augment\_\_brightness\_change & 58.9 {\footnotesize \textcolor{gray}{[56.0, 61.8]}} & 42.6 {\footnotesize \textcolor{gray}{[39.6, 45.5]}} & 16.3\\
se\_resnet50\_imagenet\_pretrained & 58.8 {\footnotesize \textcolor{gray}{[55.9, 61.7]}} & 44.1 {\footnotesize \textcolor{gray}{[41.1, 47.1]}} & 14.7\\
se\_resnet152\_imagenet\_pretrained & 58.8 {\footnotesize \textcolor{gray}{[55.9, 61.7]}} & 44.8 {\footnotesize \textcolor{gray}{[41.9, 47.8]}} & 14.0\\
cafferesnet101\_imagenet\_pretrained & 58.2 {\footnotesize \textcolor{gray}{[55.2, 61.1]}} & 44.3 {\footnotesize \textcolor{gray}{[41.3, 47.3]}} & 13.9\\
resnet50\_augment\_\_regular & 58.0 {\footnotesize \textcolor{gray}{[55.1, 61.0]}} & 42.9 {\footnotesize \textcolor{gray}{[39.9, 45.8]}} & 15.1\\
resnet34\_imagenet\_pretrained & 57.9 {\footnotesize \textcolor{gray}{[55.0, 60.9]}} & 42.8 {\footnotesize \textcolor{gray}{[39.8, 45.7]}} & 15.1\\
vgg19\_imagenet\_pretrained & 57.5 {\footnotesize \textcolor{gray}{[54.6, 60.5]}} & 40.1 {\footnotesize \textcolor{gray}{[37.2, 43.1]}} & 17.4\\
resnet50\_augment\_\_gaussian\_blur & 57.5 {\footnotesize \textcolor{gray}{[54.5, 60.4]}} & 41.8 {\footnotesize \textcolor{gray}{[38.9, 44.7]}} & 15.7\\
vgg16\_bn\_imagenet\_pretrained & 57.2 {\footnotesize \textcolor{gray}{[54.2, 60.1]}} & 39.6 {\footnotesize \textcolor{gray}{[36.7, 42.6]}} & 17.6\\
resnet50\_imagenet\_pretrained & 57.0 {\footnotesize \textcolor{gray}{[54.1, 60.0]}} & 43.8 {\footnotesize \textcolor{gray}{[40.9, 46.8]}} & 13.2\\
vgg19\_bn\_imagenet\_pretrained & 56.8 {\footnotesize \textcolor{gray}{[53.9, 59.8]}} & 40.6 {\footnotesize \textcolor{gray}{[37.7, 43.5]}} & 16.2\\
vgg16\_imagenet\_pretrained & 55.4 {\footnotesize \textcolor{gray}{[52.4, 58.4]}} & 40.1 {\footnotesize \textcolor{gray}{[37.2, 43.1]}} & 15.3\\
vgg13\_bn\_imagenet\_pretrained & 54.8 {\footnotesize \textcolor{gray}{[51.8, 57.7]}} & 38.6 {\footnotesize \textcolor{gray}{[35.7, 41.6]}} & 16.2\\
vgg11\_bn\_imagenet\_pretrained & 54.8 {\footnotesize \textcolor{gray}{[51.8, 57.7]}} & 38.8 {\footnotesize \textcolor{gray}{[35.9, 41.8]}} & 16.0\\
vgg11\_imagenet\_pretrained & 54.7 {\footnotesize \textcolor{gray}{[51.7, 57.6]}} & 38.4 {\footnotesize \textcolor{gray}{[35.5, 41.3]}} & 16.3\\
resnet18\_imagenet\_pretrained & 54.4 {\footnotesize \textcolor{gray}{[51.4, 57.4]}} & 38.1 {\footnotesize \textcolor{gray}{[35.2, 41.0]}} & 16.3\\
vgg13\_imagenet\_pretrained & 54.2 {\footnotesize \textcolor{gray}{[51.3, 57.2]}} & 37.7 {\footnotesize \textcolor{gray}{[34.9, 40.7]}} & 16.5\\
ResNeXtDenoiseAll-101\_robust\_pgd & 53.6 {\footnotesize \textcolor{gray}{[50.7, 56.6]}} & 43.2 {\footnotesize \textcolor{gray}{[40.2, 46.1]}} & 10.4\\
squeezenet1\_0\_imagenet\_pretrained & 51.1 {\footnotesize \textcolor{gray}{[48.1, 54.1]}} & 33.1 {\footnotesize \textcolor{gray}{[30.3, 36.0]}} & 18.0\\
squeezenet1\_1\_imagenet\_pretrained & 48.6 {\footnotesize \textcolor{gray}{[45.6, 51.6]}} & 31.3 {\footnotesize \textcolor{gray}{[28.6, 34.2]}} & 17.3\\
resnet50\_augment\_\_defocus\_blur & 48.4 {\footnotesize \textcolor{gray}{[45.4, 51.4]}} & 29.1 {\footnotesize \textcolor{gray}{[26.4, 31.8]}} & 19.3\\
alexnet\_imagenet\_pretrained & 45.3 {\footnotesize \textcolor{gray}{[42.4, 48.3]}} & 30.5 {\footnotesize \textcolor{gray}{[27.8, 33.3]}} & 14.8\\
    \bottomrule 
    \hiderowcolors
    \caption{\label{apdx:full_clf_results_ytbb} Classification model perturbed and original accuracies for all models in our test bed evaluated on the YTBB-robust dataset..}
                                                \end{longtable}
    

%% file: main_arxiv.bbl
\begin{thebibliography}{1}

\bibitem{bertinetto2016fully}
Luca Bertinetto, Jack Valmadre, Joao~F Henriques, Andrea Vedaldi, and Philip~HS
  Torr.
\newblock Fully-convolutional siamese networks for object tracking.
\newblock In {\em European conference on computer vision}, pages 850--865.
  Springer, 2016.

\bibitem{cadene}
Remi Cadene.
\newblock Pretrained models for pytorch.
\newblock \url{https://github.com/Cadene/pretrained-models.pytorch}.
\newblock Accessed: 2019-05-20.

\bibitem{gu2019using}
Keren Gu, Brandon Yang, Jiquan Ngiam, Quoc Le, and Jonathan Shlens.
\newblock Using videos to evaluate image model robustness.
\newblock {\em arXiv preprint arXiv:1904.10076}, 2019.

\bibitem{han2016seq}
Wei Han, Pooya Khorrami, Tom~Le Paine, Prajit Ramachandran, Mohammad
  Babaeizadeh, Honghui Shi, Jianan Li, Shuicheng Yan, and Thomas~S Huang.
\newblock Seq-nms for video object detection.
\newblock {\em arXiv preprint arXiv:1602.08465}, 2016.

\bibitem{cocomapsite}
Tsung-Yi Lin, Michael Maire, Serge Belongie, James Hays, Pietro Perona, Deva
  Ramanan, Piotr Doll{\'a}r, and C~Lawrence Zitnick.
\newblock {MS COCO} detection evaluation.
\newblock \url{http://cocodataset.org/#detection-eval}.
\newblock Accessed: 2019-05-16.

\bibitem{massa2018mrcnn}
Francisco Massa and Ross Girshick.
\newblock {maskrcnn-benchmark: Fast, modular reference implementation of
  Instance Segmentation and Object Detection algorithms in PyTorch}.
\newblock \url{https://github.com/facebookresearch/maskrcnn-benchmark}, 2018.
\newblock Accessed: 2019-05-20.

\bibitem{xiao2018video}
Fanyi Xiao and Yong Jae~Lee.
\newblock Video object detection with an aligned spatial-temporal memory.
\newblock In {\em Proceedings of the European Conference on Computer Vision
  (ECCV)}, pages 485--501, 2018.

\end{thebibliography}


\begin{thebibliography}{32}
\providecommand{\natexlab}[1]{#1}
\providecommand{\url}[1]{\texttt{#1}}
\expandafter\ifx\csname urlstyle\endcsname\relax
  \providecommand{\doi}[1]{doi: #1}\else
  \providecommand{\doi}{doi: \begingroup \urlstyle{rm}\Url}\fi

\bibitem[Azulay and Weiss(2018)]{azulay2018deep}
Aharon Azulay and Yair Weiss.
\newblock Why do deep convolutional networks generalize so poorly to small
  image transformations?
\newblock \emph{arXiv preprint arXiv:1805.12177}, 2018.

\bibitem[Bertinetto et~al.(2016)Bertinetto, Valmadre, Henriques, Vedaldi, and
  Torr]{bertinetto2016fully}
Luca Bertinetto, Jack Valmadre, Joao~F Henriques, Andrea Vedaldi, and Philip~HS
  Torr.
\newblock Fully-convolutional siamese networks for object tracking.
\newblock In \emph{European conference on computer vision}, pages 850--865.
  Springer, 2016.

\bibitem[Biggio and Roli(2018)]{biggio2017wild}
Battista Biggio and Fabio Roli.
\newblock Wild patterns: Ten years after the rise of adversarial machine
  learning.
\newblock \emph{Pattern Recognition}, 2018.
\newblock \url{https://arxiv.org/abs/1712.03141}.

\bibitem[Cadene()]{cadene}
Remi Cadene.
\newblock Pretrained models for pytorch.
\newblock \url{https://github.com/Cadene/pretrained-models.pytorch}.
\newblock Accessed: 2019-05-20.

\bibitem[Dai et~al.(2016)Dai, Li, He, and Sun]{dai2016r}
Jifeng Dai, Yi~Li, Kaiming He, and Jian Sun.
\newblock R-fcn: Object detection via region-based fully convolutional
  networks.
\newblock In \emph{Advances in neural information processing systems}, pages
  379--387, 2016.

\bibitem[Engstrom et~al.(2017)Engstrom, Tran, Tsipras, Schmidt, and
  Madry]{engstrom2017rotation}
Logan Engstrom, Brandon Tran, Dimitris Tsipras, Ludwig Schmidt, and Aleksander
  Madry.
\newblock A rotation and a translation suffice: Fooling cnns with simple
  transformations.
\newblock \emph{arXiv preprint arXiv:1712.02779}, 2017.

\bibitem[Fawzi and Frossard(2015)]{fawzi2015manitest}
Alhussein Fawzi and Pascal Frossard.
\newblock Manitest: Are classifiers really invariant?
\newblock In \emph{British Machine Vision Conference (BMVC)}, 2015.

\bibitem[Feichtenhofer et~al.(2017)Feichtenhofer, Pinz, and
  Zisserman]{feichtenhofer2017detect}
Christoph Feichtenhofer, Axel Pinz, and Andrew Zisserman.
\newblock Detect to track and track to detect.
\newblock In \emph{Proceedings of the IEEE International Conference on Computer
  Vision}, pages 3038--3046, 2017.

\bibitem[Geirhos et~al.(2018{\natexlab{a}})Geirhos, Rubisch, Michaelis, Bethge,
  Wichmann, and Brendel]{geirhos2018imagenet}
Robert Geirhos, Patricia Rubisch, Claudio Michaelis, Matthias Bethge, Felix~A
  Wichmann, and Wieland Brendel.
\newblock Imagenet-trained cnns are biased towards texture; increasing shape
  bias improves accuracy and robustness.
\newblock \emph{arXiv preprint arXiv:1811.12231}, 2018{\natexlab{a}}.

\bibitem[Geirhos et~al.(2018{\natexlab{b}})Geirhos, Temme, Rauber, Sch{\"u}tt,
  Bethge, and Wichmann]{geirhos2018generalisation}
Robert Geirhos, Carlos~RM Temme, Jonas Rauber, Heiko~H Sch{\"u}tt, Matthias
  Bethge, and Felix~A Wichmann.
\newblock Generalisation in humans and deep neural networks.
\newblock In \emph{Advances in Neural Information Processing Systems}, pages
  7538--7550, 2018{\natexlab{b}}.

\bibitem[Goodfellow et~al.(2014)Goodfellow, Shlens, and
  Szegedy]{goodfellow2014explaining}
Ian~J Goodfellow, Jonathon Shlens, and Christian Szegedy.
\newblock Explaining and harnessing adversarial examples.
\newblock \emph{arXiv preprint arXiv:1412.6572}, 2014.

\bibitem[Gu et~al.(2019)Gu, Yang, Ngiam, Le, and Shlens]{gu2019using}
Keren Gu, Brandon Yang, Jiquan Ngiam, Quoc Le, and Jonathan Shlens.
\newblock Using videos to evaluate image model robustness.
\newblock \emph{arXiv preprint arXiv:1904.10076}, 2019.

\bibitem[Han et~al.(2016)Han, Khorrami, Paine, Ramachandran, Babaeizadeh, Shi,
  Li, Yan, and Huang]{han2016seq}
Wei Han, Pooya Khorrami, Tom~Le Paine, Prajit Ramachandran, Mohammad
  Babaeizadeh, Honghui Shi, Jianan Li, Shuicheng Yan, and Thomas~S Huang.
\newblock Seq-nms for video object detection.
\newblock \emph{arXiv preprint arXiv:1602.08465}, 2016.

\bibitem[Hendrycks and Dietterich(2019)]{hendrycks2019benchmarking}
Dan Hendrycks and Thomas Dietterich.
\newblock Benchmarking neural network robustness to common corruptions and
  perturbations.
\newblock \emph{arXiv preprint arXiv:1903.12261}, 2019.

\bibitem[Hosseini and Poovendran(2018)]{hosseini2018semantic}
Hossein Hosseini and Radha Poovendran.
\newblock Semantic adversarial examples.
\newblock In \emph{Proceedings of the IEEE Conference on Computer Vision and
  Pattern Recognition Workshops}, pages 1614--1619, 2018.

\bibitem[Jin et~al.(2018)Jin, RoyChowdhury, Jiang, Singh, Prasad, Chakraborty,
  and Learned-Miller]{jin2018unsupervised}
SouYoung Jin, Aruni RoyChowdhury, Huaizu Jiang, Ashish Singh, Aditya Prasad,
  Deep Chakraborty, and Erik Learned-Miller.
\newblock Unsupervised hard example mining from videos for improved object
  detection.
\newblock In \emph{ECCV}, 2018.

\bibitem[Kanbak et~al.(2017)Kanbak, Moosavi-Dezfooli, and
  Frossard]{kanbak2017geometric}
Can Kanbak, Seyed-Mohsen Moosavi-Dezfooli, and Pascal Frossard.
\newblock Geometric robustness of deep networks: analysis and improvement.
\newblock \emph{arXiv preprint arXiv:1711.09115}, 2017.

\bibitem[Kang et~al.(2017)Kang, Li, Xiao, Ouyang, Yan, Liu, and
  Wang]{kang2017object}
Kai Kang, Hongsheng Li, Tong Xiao, Wanli Ouyang, Junjie Yan, Xihui Liu, and
  Xiaogang Wang.
\newblock Object detection in videos with tubelet proposal networks.
\newblock In \emph{Proceedings of the IEEE Conference on Computer Vision and
  Pattern Recognition}, pages 727--735, 2017.

\bibitem[Lin et~al.()Lin, Maire, Belongie, Hays, Perona, Ramanan, Doll{\'a}r,
  and Zitnick]{cocomapsite}
Tsung-Yi Lin, Michael Maire, Serge Belongie, James Hays, Pietro Perona, Deva
  Ramanan, Piotr Doll{\'a}r, and C~Lawrence Zitnick.
\newblock {MS COCO} detection evaluation.
\newblock \url{http://cocodataset.org/#detection-eval}.
\newblock Accessed: 2019-05-16.

\bibitem[Lin et~al.(2014)Lin, Maire, Belongie, Hays, Perona, Ramanan,
  Doll{\'a}r, and Zitnick]{lin2014microsoft}
Tsung-Yi Lin, Michael Maire, Serge Belongie, James Hays, Pietro Perona, Deva
  Ramanan, Piotr Doll{\'a}r, and C~Lawrence Zitnick.
\newblock Microsoft coco: Common objects in context.
\newblock In \emph{European conference on computer vision}, pages 740--755.
  Springer, 2014.

\bibitem[Massa and Girshick(2018)]{massa2018mrcnn}
Francisco Massa and Ross Girshick.
\newblock {maskrcnn-benchmark: Fast, modular reference implementation of
  Instance Segmentation and Object Detection algorithms in PyTorch}.
\newblock \url{https://github.com/facebookresearch/maskrcnn-benchmark}, 2018.
\newblock Accessed: 2019-05-20.

\bibitem[Miller(1995)]{wordnet}
George~A Miller.
\newblock Wordnet: a lexical database for english.
\newblock \emph{Communications of the ACM}, 38\penalty0 (11):\penalty0 39--41,
  1995.

\bibitem[Pashler(1988)]{pashler1988familiarity}
Harold Pashler.
\newblock Familiarity and visual change detection.
\newblock \emph{Perception \& psychophysics}, 44\penalty0 (4):\penalty0
  369--378, 1988.

\bibitem[Real et~al.(2017)Real, Shlens, Mazzocchi, Pan, and
  Vanhoucke]{real2017youtube}
Esteban Real, Jonathon Shlens, Stefano Mazzocchi, Xin Pan, and Vincent
  Vanhoucke.
\newblock Youtube-boundingboxes: A large high-precision human-annotated data
  set for object detection in video.
\newblock In \emph{Proceedings of the IEEE Conference on Computer Vision and
  Pattern Recognition}, pages 5296--5305, 2017.

\bibitem[Recht et~al.(2019)Recht, Roelofs, Schmidt, and
  Shankar]{recht2019imagenet}
Benjamin Recht, Rebecca Roelofs, Ludwig Schmidt, and Vaishaal Shankar.
\newblock Do imagenet classifiers generalize to imagenet?
\newblock \emph{arXiv preprint arXiv:1902.10811}, 2019.

\bibitem[Ren et~al.(2015)Ren, He, Girshick, and Sun]{ren2015faster}
Shaoqing Ren, Kaiming He, Ross Girshick, and Jian Sun.
\newblock Faster r-cnn: Towards real-time object detection with region proposal
  networks.
\newblock In \emph{Advances in neural information processing systems}, pages
  91--99, 2015.

\bibitem[Russakovsky et~al.(2015)Russakovsky, Deng, Su, Krause, Satheesh, Ma,
  Huang, Karpathy, Khosla, Bernstein, Berg, and Fei-Fei]{ILSVRC15}
Olga Russakovsky, Jia Deng, Hao Su, Jonathan Krause, Sanjeev Satheesh, Sean Ma,
  Zhiheng Huang, Andrej Karpathy, Aditya Khosla, Michael Bernstein,
  Alexander~C. Berg, and Li~Fei-Fei.
\newblock {ImageNet Large Scale Visual Recognition Challenge}.
\newblock \emph{IJCV}, 115\penalty0 (3):\penalty0 211--252, 2015.
\newblock \doi{10.1007/s11263-015-0816-y}.

\bibitem[Torralba et~al.(2011)Torralba, Efros, et~al.]{torralba2011unbiased}
Antonio Torralba, Alexei~A Efros, et~al.
\newblock Unbiased look at dataset bias.
\newblock In \emph{CVPR}, volume~1, page~7. Citeseer, 2011.

\bibitem[Xiao and Jae~Lee(2018)]{xiao2018video}
Fanyi Xiao and Yong Jae~Lee.
\newblock Video object detection with an aligned spatial-temporal memory.
\newblock In \emph{Proceedings of the European Conference on Computer Vision
  (ECCV)}, pages 485--501, 2018.

\bibitem[Xie et~al.(2018)Xie, Wu, van~der Maaten, Yuille, and
  He]{xie2018feature}
Cihang Xie, Yuxin Wu, Laurens van~der Maaten, Alan Yuille, and Kaiming He.
\newblock Feature denoising for improving adversarial robustness.
\newblock \emph{arXiv preprint arXiv:1812.03411}, 2018.

\bibitem[Zheng et~al.(2016)Zheng, Song, Leung, and Goodfellow]{Zheng_2016}
Stephan Zheng, Yang Song, Thomas Leung, and Ian Goodfellow.
\newblock Improving the robustness of deep neural networks via stability
  training.
\newblock \emph{2016 IEEE Conference on Computer Vision and Pattern Recognition
  (CVPR)}, Jun 2016.
\newblock \doi{10.1109/cvpr.2016.485}.
\newblock URL \url{http://dx.doi.org/10.1109/cvpr.2016.485}.

\bibitem[Zhu et~al.(2017)Zhu, Wang, Dai, Yuan, and Wei]{zhu2017flow}
Xizhou Zhu, Yujie Wang, Jifeng Dai, Lu~Yuan, and Yichen Wei.
\newblock Flow-guided feature aggregation for video object detection.
\newblock In \emph{Proceedings of the IEEE International Conference on Computer
  Vision}, pages 408--417, 2017.

\end{thebibliography}
